\documentclass[10pt,journal,compsoc]{IEEEtran}
\usepackage{graphicx}
\usepackage{amsmath,amssymb} 
\usepackage{times}
\usepackage{epsfig}
\usepackage{graphicx}
\usepackage{algorithm}
\usepackage[noend]{algpseudocode}
\usepackage{multirow}
\usepackage{array}
\usepackage{subfigure}
\usepackage{mathtools}
\usepackage{amsmath}
\usepackage{tabularx}
\usepackage{wasysym}
\usepackage{url}
\usepackage{hyperref}

\begin{document}
\title{Latent-Class Hough Forests for 6~DoF Object Pose Estimation}

\author{Rigas~Kouskouridas, Alykhan~Tejani, Andreas~Doumanoglou, Danhang~Tang, Tae-Kyun~Kim,~\IEEEmembership{Member ~IEEE}
\IEEEcompsocitemizethanks{
\IEEEcompsocthanksitem R. Kouskouridas and A. Tejani contributed equally to this work.

\IEEEcompsocthanksitem R. Kouskouridas, A. Doumanoglou, D. Tang and TK Kim are with the Department of Electrical and Electronic Engineering, Imperial College London, SW7 2AZ, UK. E-mail: \{r.kouskouridas, a.doumanoglou12, d.tang11, tk.kim\}@imperial.ac.uk
\IEEEcompsocthanksitem A. Tejani is with Blippar, 1 London Bridge, London, SE1, UK. E-mail: info@alykhantejani.com}}



\IEEEtitleabstractindextext{%
\begin{abstract}
In this paper we present \textit{Latent-Class Hough Forests}, a method for object detection and 6 DoF pose estimation in heavily cluttered and occluded scenarios. We adapt a state of the art template matching feature into a scale-invariant patch descriptor and integrate it into a regression forest using a novel template-based split function. We train with positive samples only and we treat class distributions at the leaf nodes as latent variables. During testing we infer by iteratively updating these distributions, providing accurate estimation of background clutter and foreground occlusions and, thus, better detection rate. Furthermore, as a by-product, our \textit{Latent-Class Hough Forests} can provide accurate occlusion aware segmentation masks, even in the multi-instance scenario. In addition to an existing public dataset, which contains only single-instance sequences with large amounts of clutter, we have collected two, more challenging, datasets for multiple-instance detection containing heavy 2D and 3D clutter as well as foreground occlusions. We provide extensive experiments on the various parameters of the framework such as patch size, number of trees and number of iterations to infer class distributions at test time. We also evaluate the \textit{Latent-Class Hough Forests} on all datasets where we outperform state of the art methods.
\end{abstract}

}

\maketitle
\IEEEdisplaynontitleabstractindextext
\IEEEpeerreviewmaketitle


\newcolumntype{L}[1]{>{\raggedright\arraybackslash}p{#1}}
\newcolumntype{C}[1]{>{\centering\arraybackslash}p{#1}}
\newcolumntype{R}[1]{>{\raggedleft\arraybackslash}p{#1}}

\newcommand{\vect}[1]{\boldsymbol{\mathbf{#1}}}

\newcommand{\regressionforest}{\mathcal{F}}

\newcommand{\splitfunctions}{\mathcal{F}}
\newcommand{\splitfunction}{f}
\newcommand{\splitcandidates}{\Phi}
\newcommand{\splitcandidate}{\phi}
\newcommand{\totalset}{\mathcal{S}}
\newcommand{\depthdata}{d}
\newcommand{\labeldata}{l}
\newcommand{\offsetu}{u}
\newcommand{\offsetv}{v}
\newcommand{\predictedoffset}{\phi \prime}
\newcommand{\groundtruthoffset}{\phi}
\newcommand{\erroroffset}{\Delta \phi}
\newcommand{\avg}{avg}
\newcommand{\ksynset}{\mathcal{K}}
\newcommand{\trvar}{\mathrm{\Lambda}}
\newcommand{\trace}{\mathrm{trace}}
\newcommand{\etal}{\emph{et al.}}

\newcommand{\assoc}{\mathrm{\Psi}}

\newcommand{\LRF}{Latent Regression Forest}
\newcommand{\LRFabbrv}{LRF}
\newcommand{\LRT}{Latent Regression Tree}
\newcommand{\LRTabbrv}{LRT}

\newcommand{\ori}{\mathbf{ori}}

\newcommand{\testimg}{\mathbf{I}} 
\newcommand{\testpatchset}{\hat{\patchset}}
\newcommand{\testpatch}{\hat{\patch}}
\newcommand{\testviewlabel}{\hat{\viewlabel}}
\newcommand{\testjointlabel}{\hat{\jointlabel}}
\newcommand{\testvotelabel}{\hat{\votelabel}}
\newcommand{\testjoint}{\mathbf{Y}}
\newcommand{\testonejoint}{\mathbf{y}}
\newcommand{\testjointset}{\mathcal{Y}}
\newcommand{\testqualthres}{t_{q}}
\newcommand{\testoccset}{\mathcal{O}}

\newcommand{\ntrainingdata}{1000}

\newcommand{\argmin}[1]{\underset{#1}{\operatorname{arg}\!\operatorname{min}}\,}

\section{Introduction}
\IEEEPARstart{A}{mong} the most challenging tasks in computer vision is the one of estimating the 3D pose of an object due to its practical implication and its fundamental importance applications like robotic manipulation~\cite{kouskouridas2014sparse} and tracking~\cite{crivellaro2015novel}. In order to efficiently fulfill the 3D object detection and pose estimation task, a computer vision method should tackle several cascading issues that hinder its effective application. Although recent emergence of consumer depth sensors provides additional cue in favour of textureless objects, background clutter, partial occlusions and large scale changes still put barriers to this problem. Template matching techniques ~\cite{hinterstoisser2013model} can tackle clutter and occlusion to some degree, but have inherent weakness due to their holistic nature. Point-to-Point approaches fail in cases of planar, self-similar or similar to background clutter objects~\cite{bariya2010scale,drost2010model} due to the fact that similar point features vote for different pose parameters. Moreover, these methods were only evaluated with the assumption of only one instance existing in the scene. The case of multiple object instances, i.e., challenging precision-recall, is left unexplored. 

Another important perspect is that, prior arts in 3D object pose estimation~\cite{bariya2010scale,drost2010model,hinterstoisser2013model} utilize mesh models of target objects to generate training samples. This implies that only positive samples are used and, thus, falls into the category of one-class learning. On the other hand, to explicitly tackle the aforementioned challenges, a more traditional way in 2D detection is to augment negative samples (clutter and occluder) during training. Due to the complexity of negative scenarios, this often results in huge amount of training samples, thus, increasing the computation burden of the system. And yet no such work can guarantee covering all cases. At present, there is a big disparity in the number of depth image datasets vs. 2D image datasets, adding a further challenging in mining for negative depth samples.

\begin{figure}
\begin{center}
\includegraphics[width=0.32\linewidth]{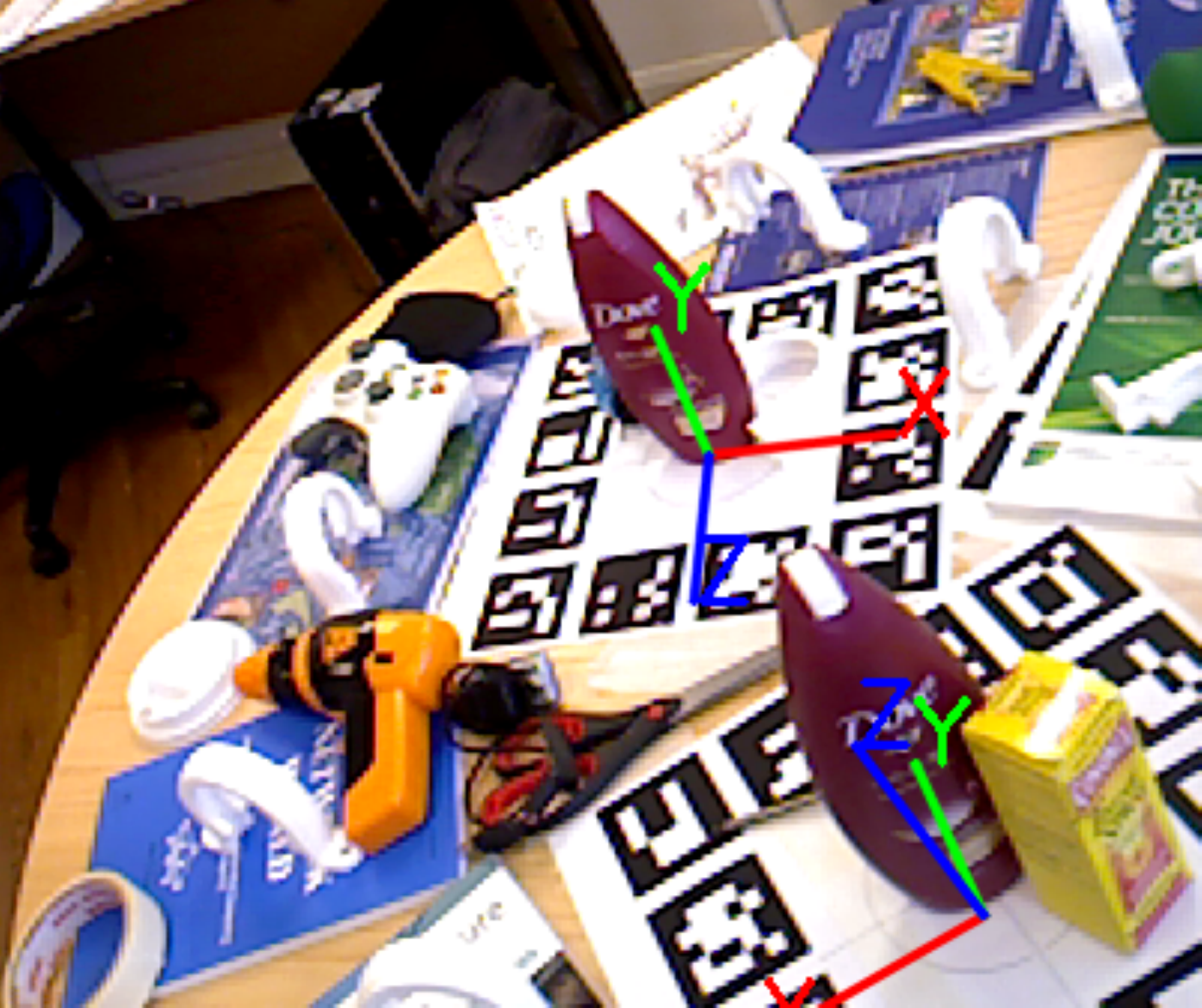}
\includegraphics[width=0.32\linewidth]{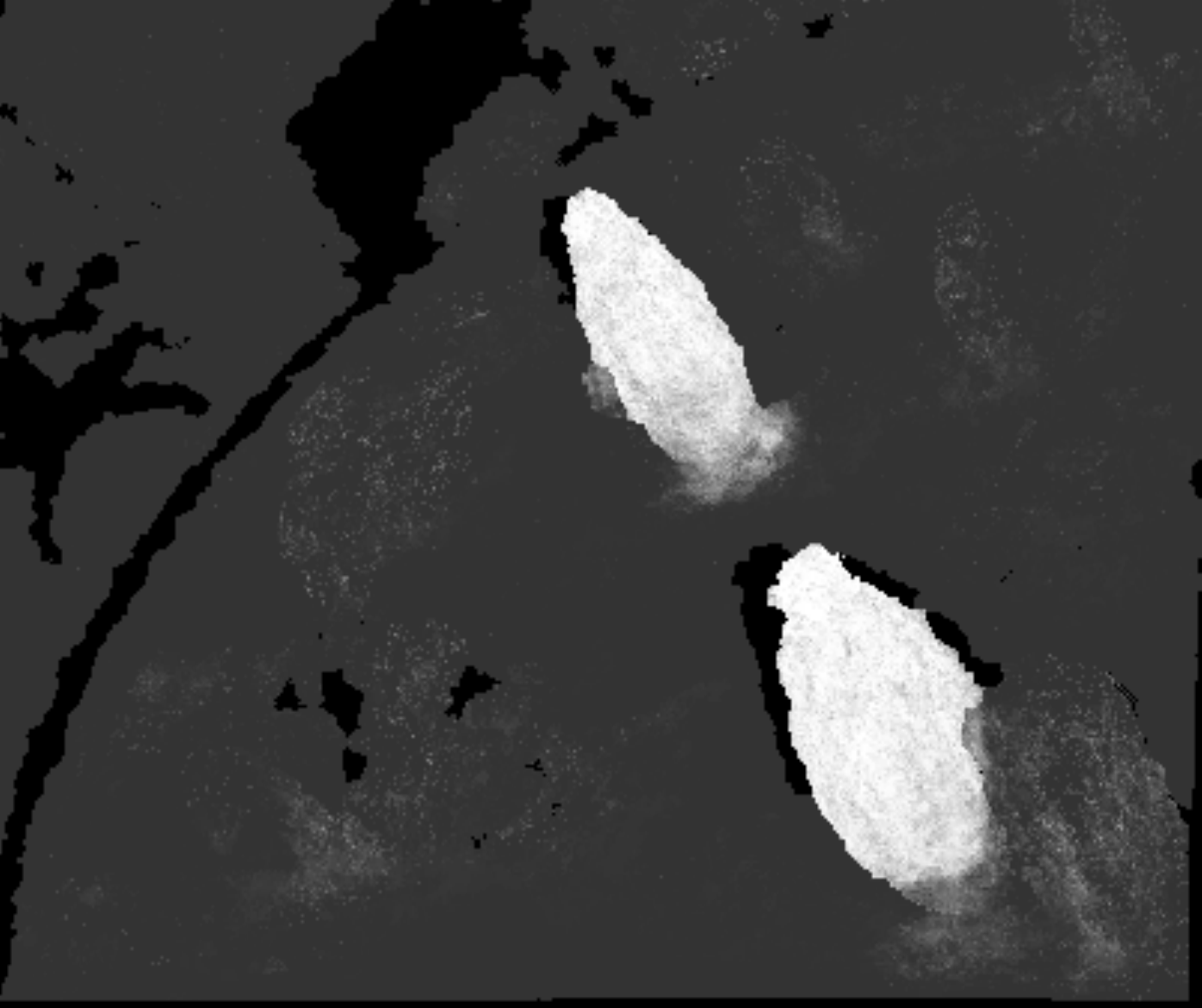}
\includegraphics[width=0.32\linewidth]{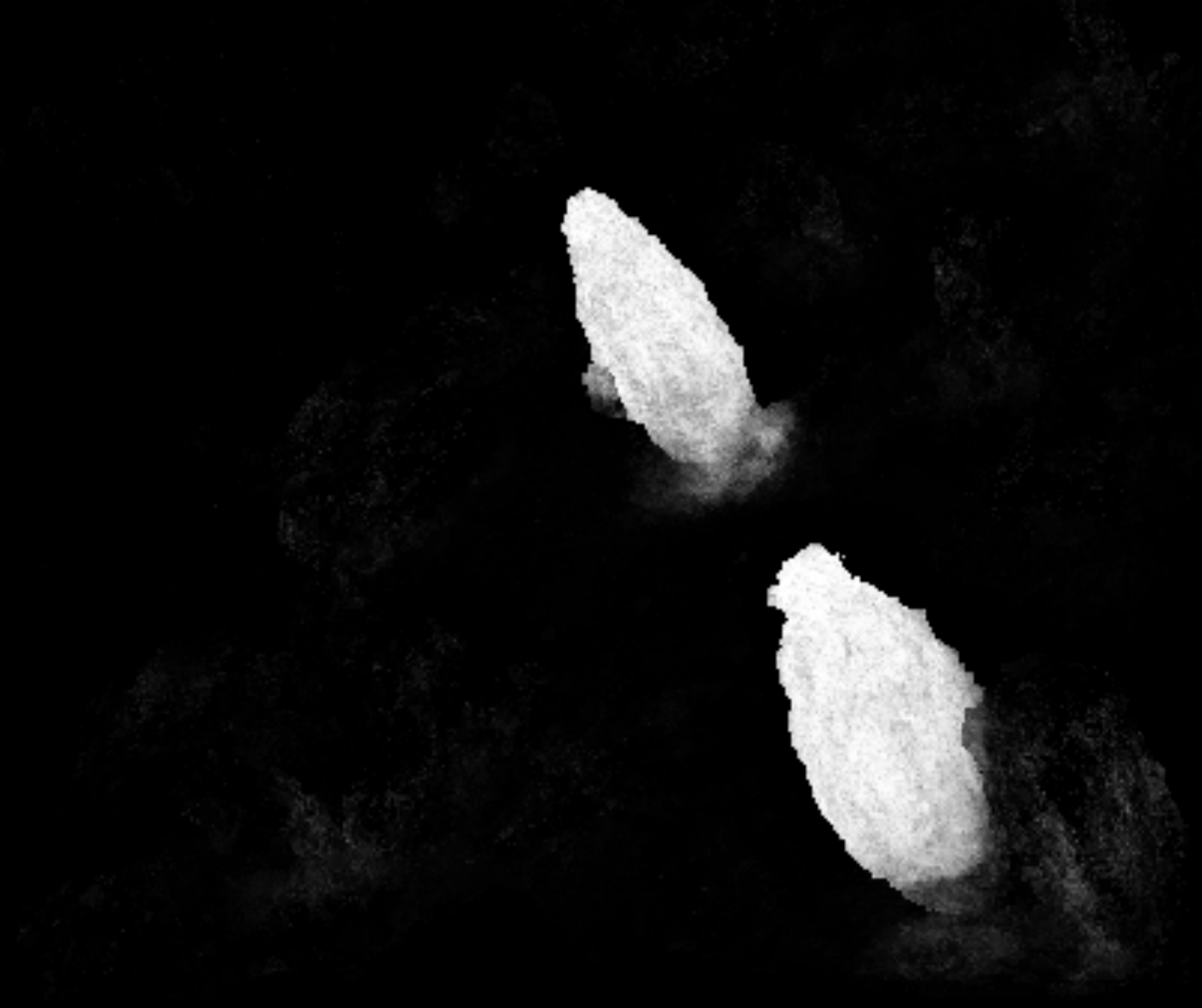}
\includegraphics[width=0.32\linewidth]{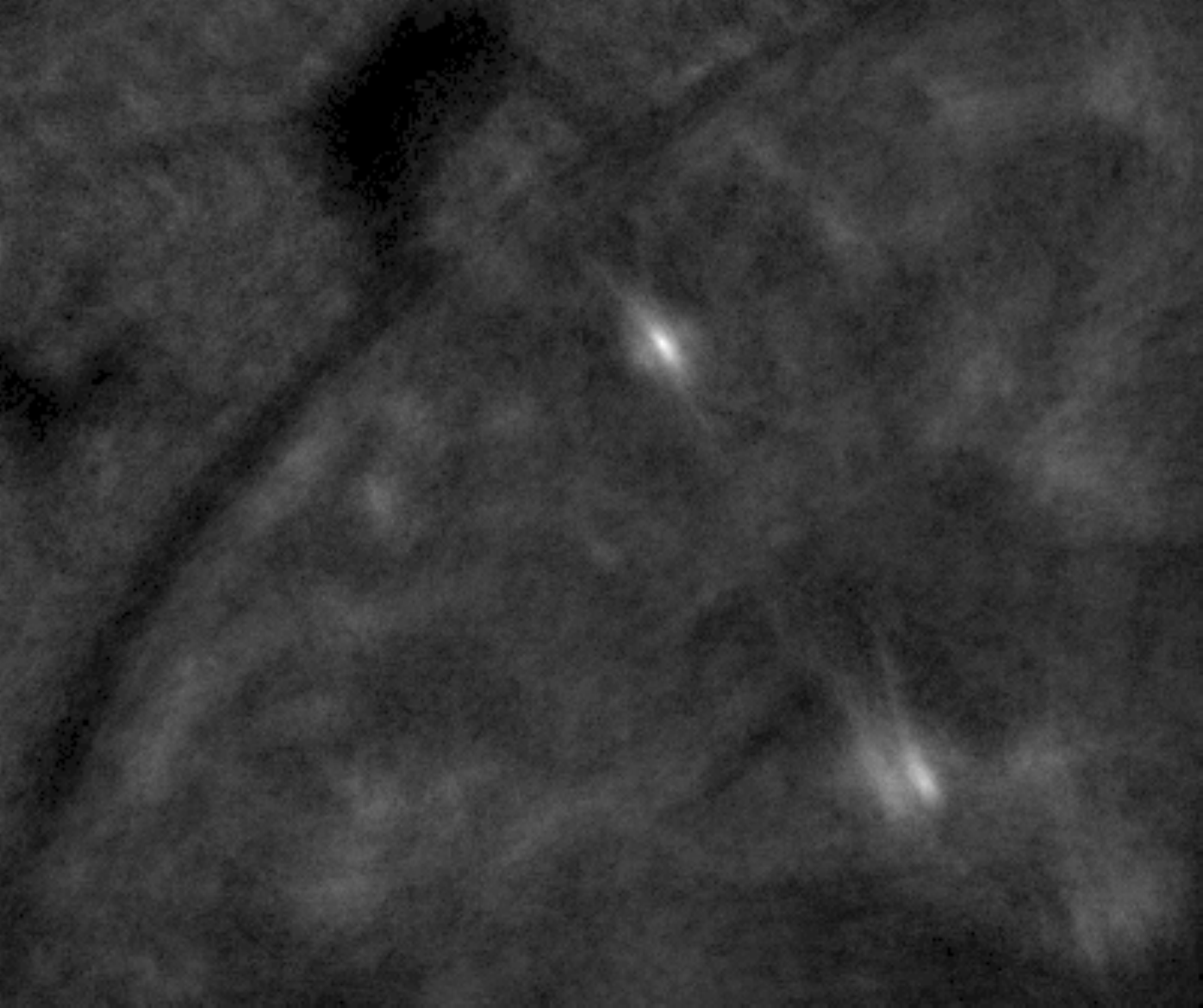}
\includegraphics[width=0.32\linewidth]{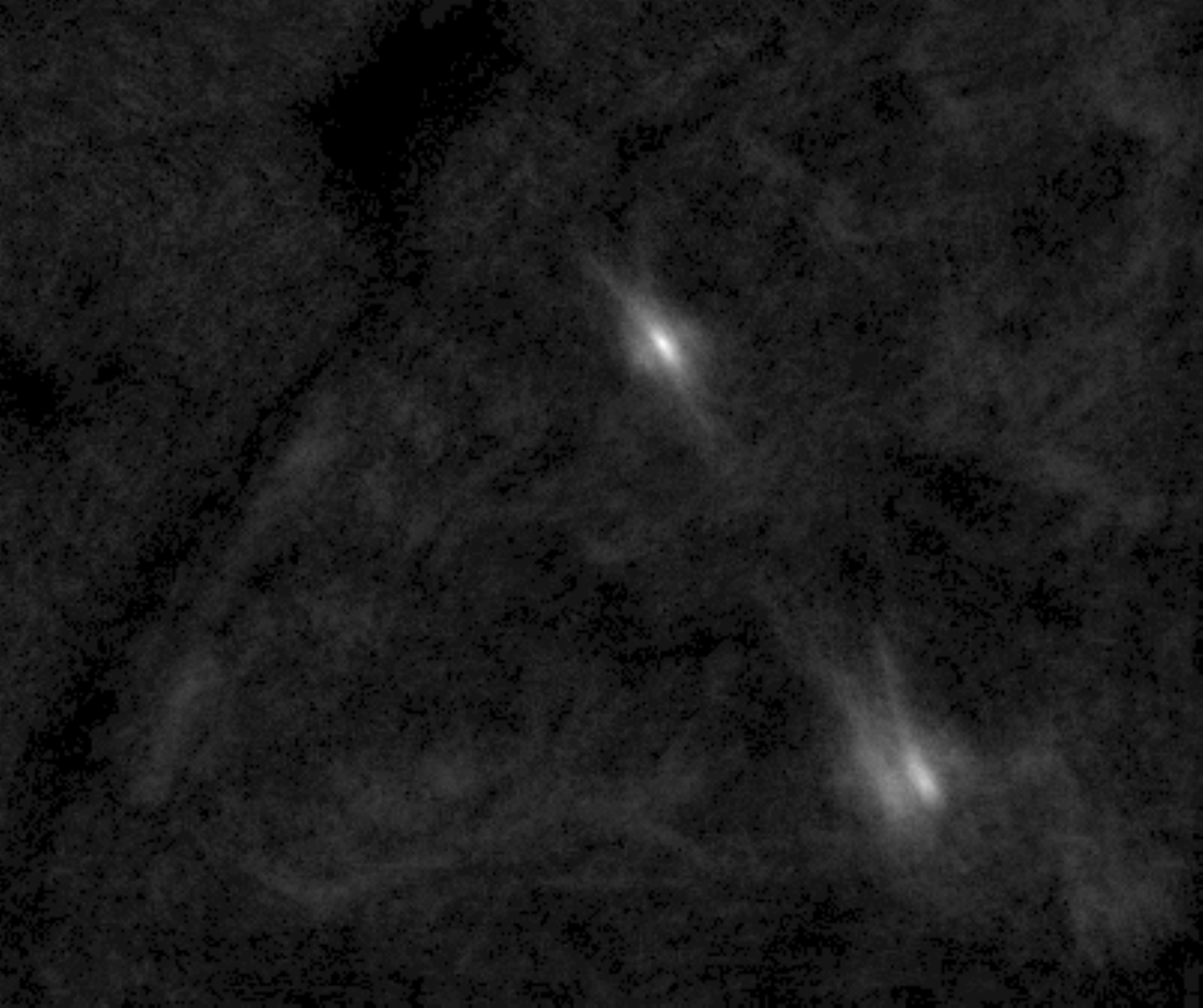}
\includegraphics[width=0.32\linewidth]{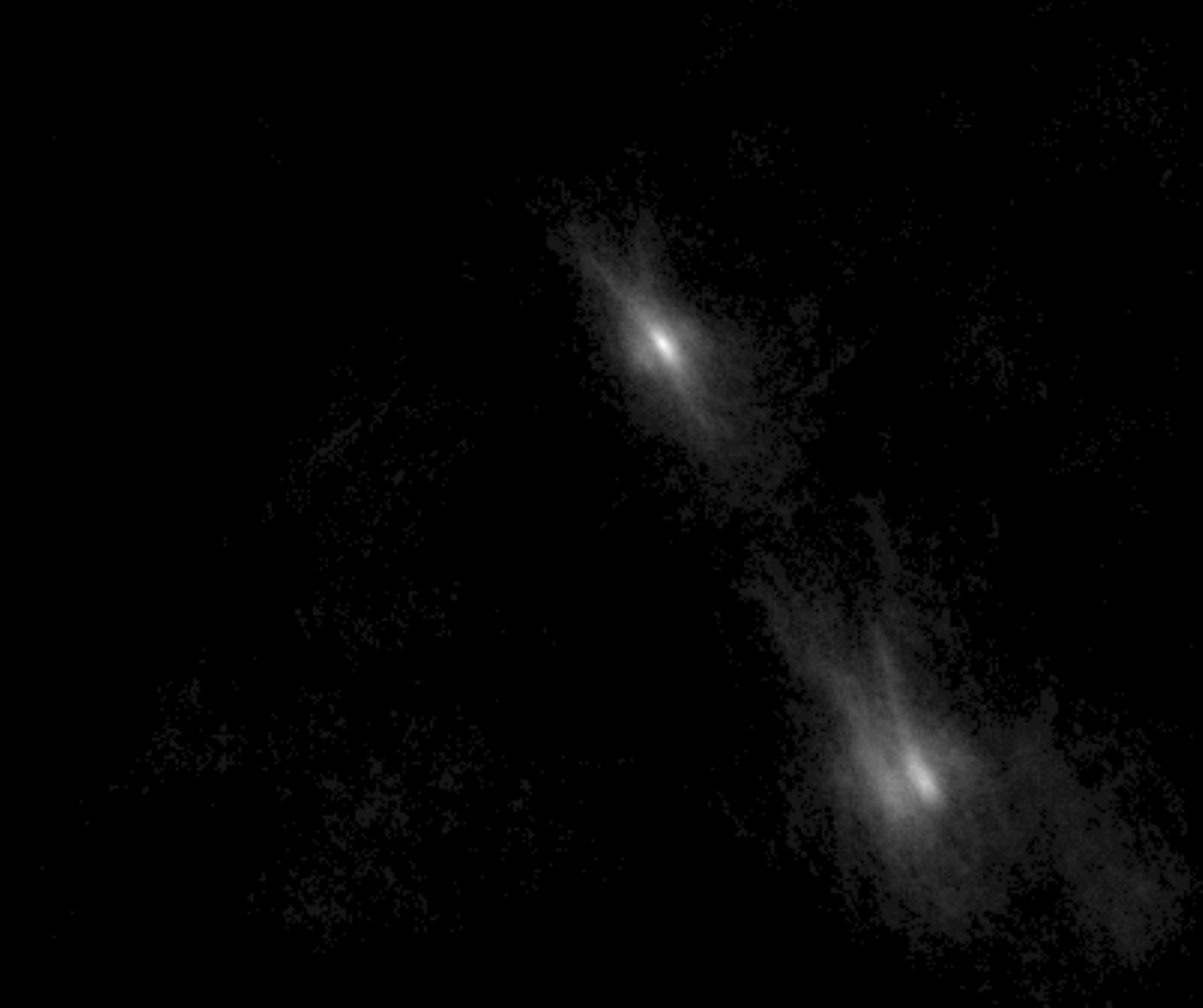}
\caption{An illustration of intermediate results of the iterative process. Column 1: Initial pose estimation and the corresponding hough map. Columns 2 - 3: Foreground probability masks and the respective hough maps after $\#5$ and $\#10$ iterations.}
\label{fig:inference_results}
\end{center}
\end{figure}

Highly motivated by these challenges, we present a novel method, called \textit{Latent-Class Hough Forests}, for 3D object detection and pose estimation. Unlike traditional Hough Forest~\cite{gall2011hough}, which explicitly exploits class information, our method utilizes only the regression term during the training stage. Also differing from a regression forest, \textit{Latent-Class Hough Forests} take into account class distribution at leaf nodes. During testing, the distribution of positive / negative classes are considered as latent variables being updated iteratively, in order to provide more and more accurate voting results. As a byproduct, our system can also produce accurate occlusion-aware segmentation masks. Figure~\ref{fig:inference_results} demonstrates the effect of our inference algorithm. At iteration\#0, the forest has no background information, thus all pixels are considered as foreground, which results in a noisy vote map. As the iterative process goes on, it is evident that background pixels are greatly suppressed and both the occlusion mask and vote map become more accurate and clean. 

The paper in hand presents a novel method for object detection and 6 DoF pose estimation in real scenarios by adopting a part-based strategy into the random forest framework. The previous conference version \cite{tejani2014latent} has been extended by a more detailed discussion of the key ingredients of the method and an extensive analysis of the parameters of the technique. Moreover, we propose an additional dataset inspired by industrial settings as well as reporting more experiments on three different datasets. Our main contributions can be summarized as follows:
\begin{itemize}
	\item We propose the \textit{Latent-Class Hough Forests}, a novel patch-based approach to 3D object detection and pose estimation; It performs one-class learning at the training stage, and iteratively infers latent class distributions at test time.
	\item We adapt the state of the art 3D holistic template feature, LINEMOD \cite{hinterstoisser2011multimodal}, to be a scale invariant patch descriptor and integrate it into the random forest framework via a novel template-based splitting function.
	\item During the inference stage, we jointly estimate objects' 3D location and pose as well as a pixel wise visibility map, which can be used as an occlusion aware figure-ground segmentation.
	\item We provide two new, more challenging public datasets for \emph{multi-instance} 3D object detection and pose estimation, comprising \emph{near and far range 2D and 3D clutter} as well as \emph{foreground occlusions} in domestic and industrial scenarios. To the best of our knowledge, we are the first to provide a fully-annotated bin-picking dataset.
\end{itemize}

In the remainder of this paper we first discuss related work in Sec. \ref{sec:related_work} before introducing our method in Sec. \ref{sec:methodology}. Following this, in Sec. \ref{sec:experiments}, we provide a quantitative and qualitative analysis of our results as well as a comparison to current state of the art methods. Finally, in Sec. \ref{sec:conclusion}, we conclude with some final remarks and a discussion of future work.


\section{Related Work}
\label{sec:related_work}
Throughout the years several techniques for the detection and registration of objects in 3D environments have been proposed. According to the literature, three main categories can be distinguished: Template matching, learning-based methods and Point-to-Point techniques. The simplicity along with facile training sessions render template matching methods as one of the most widely used solutions for object detection tasks. From the very known baseline techniques of LINEMOD~\cite{hinterstoisser2011multimodal} and its extension~\cite{Rios_Cabrera_2013_ICCV}, to the classic implementation of Distance Transform approaches~\cite{liu2012}, template matching methods have found application in contemporary vision tasks and robotics modules, respectively. On the other hand, learning-based approaches impose upon laborious training sessions with numerous training samples with view to extract highly representative object models~\cite{malisiewicz2011,kouskouridas2013efficient}. Point-to-Point techniques build upon point pair features to construct object models based on point clouds. A representative method of this category is the one presented by Drost \etal~\cite{drost2010model}. In turn, simple pixel-based features have been also employed to tackle the object pose estimation problem. More recently, Brachmann \etal~\cite{brachmann2014learning} introduced a new representation in form of a joint 3D object coordinate and class labelling (extended for tracking in \cite{krull6}), which, however, suffers in cases of occlusions. 

Moreover, in \cite{lim2014fpm} a method for fine pose estimation by representing geometric and appearance information as a collection of 3D shared parts and objectness, has been presented. Song \etal~\cite{song2014sliding} proposed a computationally expensive approach to the 6 DoF pose estimation problem that slides exemplar SVMs in the 3D space, while in \cite{bonde2014robust} shape priors are learnt by soft labelling random forest for 3D object classification and pose estimation. Wu \etal~\cite{3dShap_Net} designed a model that learns the joint distribution of voxel data and category labels using a Convolutional Deep Belief Network, while the posterior distribution for classification is approximated by Gibbs sampling. From the relevant literature we could also identify the works of Aldoma \etal~\cite{verification2012} and Buch \etal~\cite{buch2014search} that propose a final step of fine pose refinement for false positive / outlier removal. Last, Wohlhart \etal~\cite{wohlhart2015test} showed how a Convolutional Neural Network can be trained to learn a 3D pose estimation-wise descriptor. 

It is well understood that modeling objects as a collection of parts increases robustness to intra-class variation, pose change and even occlusion. The implicit shape model, introduced by Leibe \etal~\cite{leibe2004combined}, learns, via unsupervised clustering, class-specific visual codebooks and spacial distributions for each entry. Codebook entries are then detected in the test image and used to cast probabilistic votes in the Hough space based on the learnt spatial distributions. Gall and Lempitsky showed, with the class-specific Hough forest~\cite{gall2011hough}, how part-based modeling can be effectively combined with generalized Hough voting for object detection under the random forest framework~\cite{breiman2001random}. Tang \etal~\cite{tang2012fast} combined Hough Forest with DOT~\cite{hinterstoisser2010dominant} with a template matching split function, while highly efficient, requires extensive and diverse background images for training.

On the other hand, one-class training stands for the learning process that imposes upon training without negative samples. Introduced by Moya \etal~\cite{moya1996,moya1993one} and further developed by Tax~\cite{tax2001one} and Scholkopf~\cite{scholkopf2001estimating}, these approaches lay their foundations within the support vector framework and aim to derive an enclosing decision boundary of the training data as a whole from a few supporting samples. Other techniques such as the works of Bishop~\cite{bishop1994} and Para \etal~\cite{parra1996} approached the problem in a probabilistic manner and tried to find the underlying density model for the training data as a whole.

Occlusion-handling is well-related to object segmentation problem. In traditional 2D scenarios, existing methods can be categorised into tackling occlusion in training~\cite{mathias2013,pepikj2013} or inference stage~\cite{hsiao2012,kokkinos2009}. Utilizing depth cues is relatively new and recently Wang~\etal~\cite{wang2013} approached occlusion reasoning by explicitly learning clutter and oclusion scenarios during the training session, whilst our method falls into the inference category.

\section{Proposed Method}
\label{sec:methodology}

In the field of object detection and 3D pose estimation, LINEMOD~\cite{hinterstoisser2011multimodal}, a binary RGB-D feature, has demonstrated both state of the art accuracy and efficiency. However, so far it has been combined with a holistic template matching scheme, which has inherent problem with occlusion due to the nature of holism. Moreover, as a near-neighbour search, this scheme slows down linearly as the number of templates grows. And the fact that LINEMOD is not scale invariant often leads to thousands of templates per object, in order to cover multiple scales and numerous viewpoints. To be more robust to occlusion and clutter, we start off by combining a state of the art 3D feature and a part-based detector. In this work, we choose the state of the art part-based detector Hough Forest~\cite{gall2011hough}. However in our case, naively combining them does not work because: a) As a binary feature, LINEMOD only considers orientation information whilst discarding magnitute. This provides efficiency but degrades the accuracy in the case of a patch-based detector. b) No negative (clutter / occlusion) information is available during training, which means the {\it classification} term in Hough Forest cannot be adopted. c) Moreover, not knowing  the boundary between positive and negative samples leads to large amount of false positives during inference.

To address these issues, we propose \textit{Latent-Class Hough Forests} to datamine useful clutter / occlusion information from inference stage and transfer the knowledge to the detector. In section~\ref{sec:learning} we describe how to modify the similarity measurement of LINEMOD and integrate it into the split function of \textit{Latent-Class Hough Forests}. Section~\ref{sec:inference} depicts the inference process that jointly and iteratively updates the latent class distribution and voting results.
\subsection{Learning}
\label{sec:learning}
\begin{figure*}[ht]
	\begin{center}
	\includegraphics[width=0.65\linewidth]{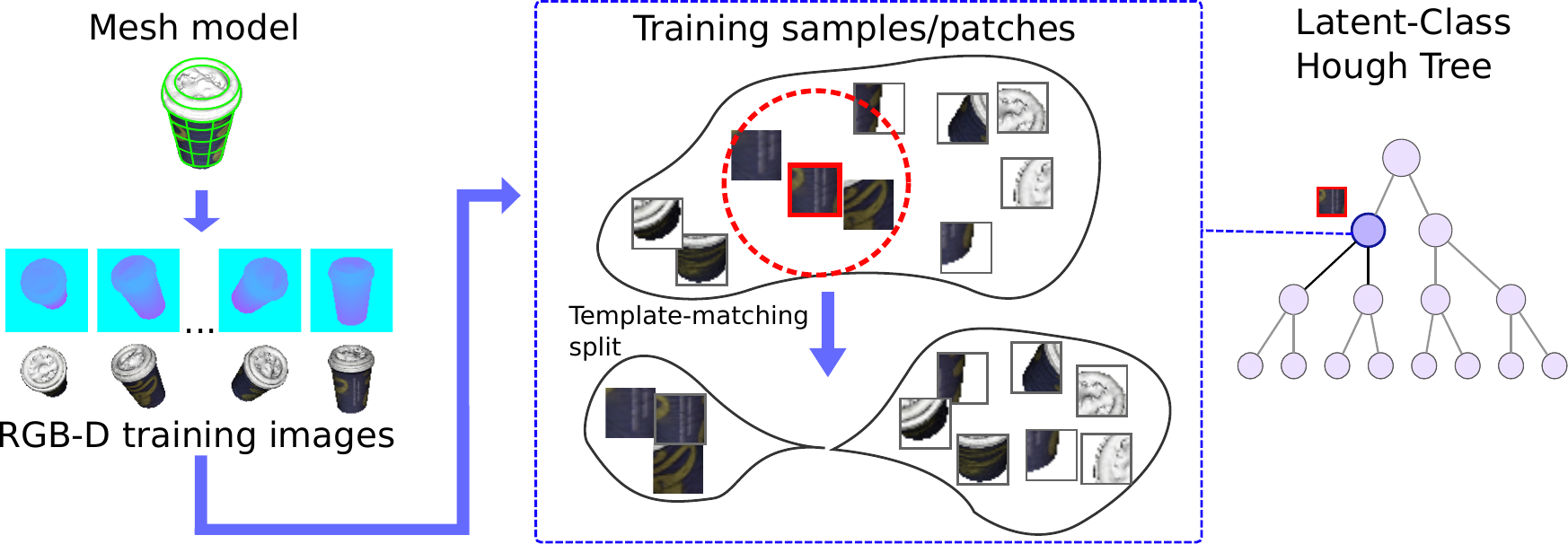}
	\caption{During training, a random patch $\mathbb{T}$ is selected (red frame) as a template. The similarity between it and all other patches is measured and splitted based on a threshold $\tau$ (dotted circle). This process is repeated until the optimal $\mathbb{T}$ and $\tau$ are found and stored in the current node.}
	\label{fig:learning}
	\end{center}
\end{figure*}
\textit{Latent-Class Hough Forests} are an ensemble of randomized binary decision trees trained using the general random forest framework \cite{breiman2001random}. During training, each tree is built using a random subset of the complete training data. Each intermediate node in the tree is assigned a split function and threshold to optimize a measure of information gain; this test is then used to route incoming samples either left or right. This process is repeated until some stopping criteria is met, where a leaf node containing application-specific contextual information is formed. Each stage in this learning process is highly application dependent and we will discuss each in turn below. Fig. \ref{fig:learning} illustrates the high level idea underlying our training module. A typical split function of a random forest can be formulated as below:
\begin{equation}
\label{eq:split}
	h_i(x) = \left\{ \begin{array}{rcl}
0, & \quad S(x, \rho_i) \leq \tau_i
\\ 1, & \quad S(x, \rho_i) > \tau_i
\end{array}\right.,
\end{equation}
where $\rho_i$ is the parameter and $\tau_i$ is the threshold stored at node $i$. $S$ is a test function that evaluates the input sample given $\rho_i$. The process of training a tree is to decide the optimal $\rho_i$ and $\tau_i$ by measuring the information gain.

\subsubsection{Training Data}
In order to capture reasonable viewpoint coverage of the target object, we render synthetic RGB and depth images by placing a virtual camera at each vertex of a subdivided icosahedron of a fixed radius. A tree is trained from a set of patches sampled from the training images. We extract patches with size relative to the bounding box of the rendered object, while the template features are evenly spread across each patch; features capturing the image gradients are taken only from the object contours and features capturing the surface normals are taken from the body of the object. Moreover, the collection and representation of template features is the same as described in \cite{hinterstoisser2011multimodal}.
 
\subsubsection{Split Function}
\begin{figure*}[ht]
	\begin{center}
	\includegraphics[width=0.65\linewidth]{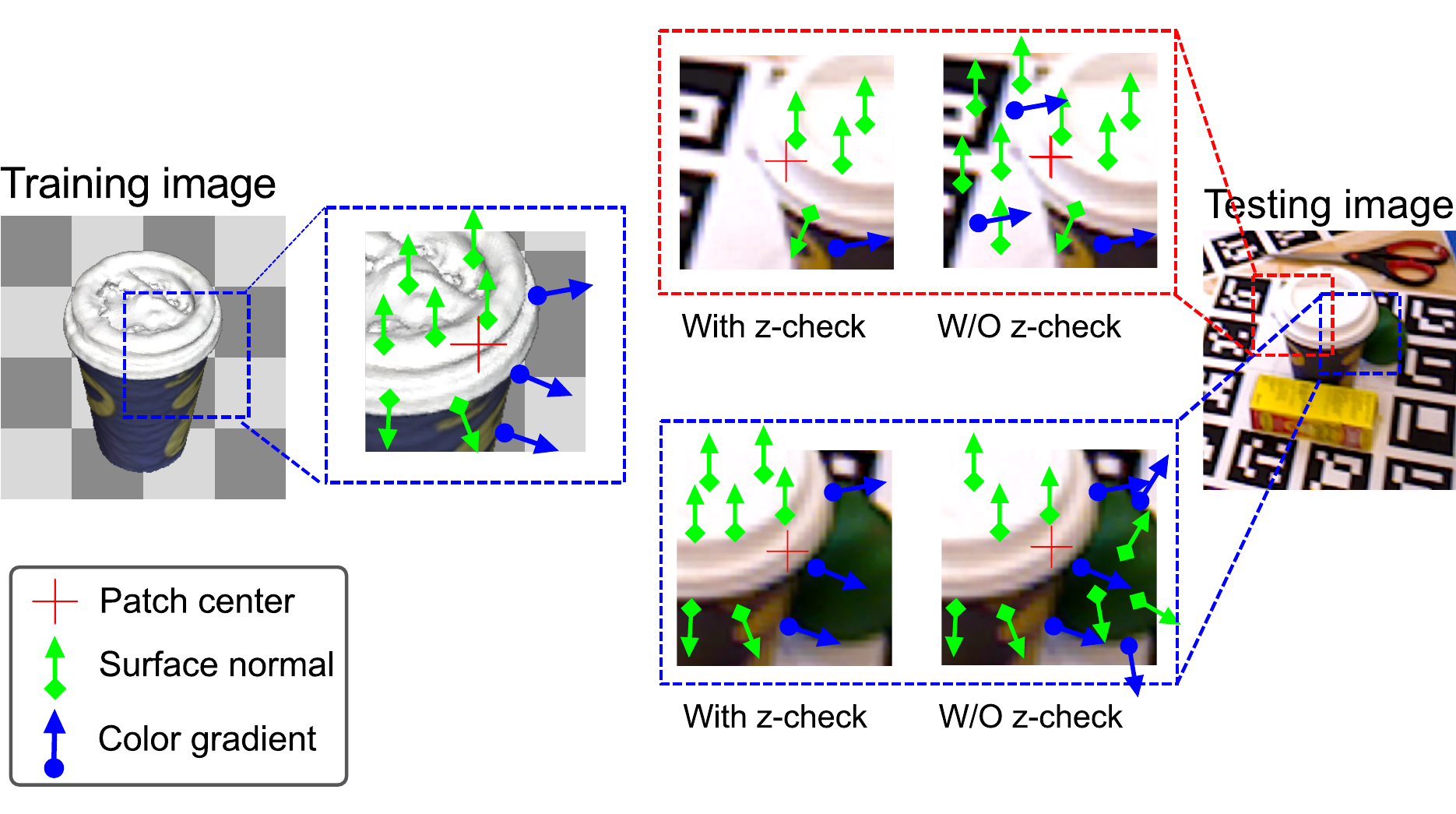}
	\caption{$z$-value check enhance the robustness against clutter and occluders. Blue patches indicate a true positive match and red ones a false positives one. Without the z-check, features of the planar background of the false patch could match the training patch and become a false positive.}
	\label{fig:feature}
	\end{center}
\end{figure*}
It has been shown in the 2D detection problem~\cite{tang2012fast} that the original test function (so called two-pixel test) of Hough Forest does not work well with a binary feature. Thus, doing a naive holistic patch comparison, or the two-dimenson / two-pixel tests (as used in \cite{shotton2013real,fanelli2011real,tangreal}) can lead to test patches taking the incorrect route at split functions. To compensate the loss of magnitude information whilst keeping the efficiency, one possible way is to utilize the orientation information in a more effective way. To this end, a non-linear template matching test function is adopted as described below:
\begin{equation}
	\label{eq:similarity}
	\begin{split}
	S(\mathcal{X}, \rho) = S(\mathcal{X}, \mathbb{T}) &= \sum \limits_{ r \in \mathcal{P}} g( \ori(\mathcal{X}, r), \ori(\mathcal{O}, r)) \\
	&=\sum \limits_{ r \in \mathcal{P}}\Big(\max_{} f_m\big(\mathcal{X}(r), \mathcal{O}(r)\big)\Big) 
	\end{split}
\end{equation}
where $\mathcal{X}$ is a testing patch, $\mathbb{T} = (\mathcal{O}, \mathcal{P})$ is the template of a patch $\mathcal{O}$ with a list $\mathcal{P}$ of features. $f_m$ is the dot product between the gradient orientation at location $r$ of RGB-D patch $\mathcal{X}$ and and $\mathcal{O}$, respectfully. The similarity measure $f_m$ is similar to \cite{hinterstoisser2011multimodal}, while, here we show how it can be adapted to work over patches. Combining Eq.~\ref{eq:split} and~\ref{eq:similarity} gives us a non-linear split function, which performs much better than axis-aligned and linear split functions, yet, has the complexity similar to an axis-aligned one, since it involves only bitwise operations that can SSE-accelerated.

The test function of Eq. \ref{eq:similarity} performs well within the object boundary, but poorly around the edge. Mainly because in inference stage, clutter and occluder around target object make the similarity measurement between patches to fail. See Fig. \ref{fig:feature} for an illustration of this issue. To tackle this, we modify the test function by incorporating an efficient $z$-value check:
\begin{equation}
\label{eq:scale_inv_similarity}
\resizebox{1\hsize}{!}{$
\begin{cases}
	S(\mathcal{X}, \mathbb{T}) &= \sum \limits_{ r \in \mathcal{P}} f(\mathcal{X}, \mathcal{O}, c, r)  g( \ori(\mathcal{X}, r), \ori(\mathcal{O}, r) ), \\
	f(\mathcal{X}, \mathcal{O}, c, r) &= \delta ( |(D(\mathcal{X}, c) - D(\mathcal{X}, r)) - (D(\mathcal{O}, c) - D(\mathcal{O}, r))| < \tau )
\end{cases}
$}
\end{equation}
where, for a patch centered at position $c$, $D(a,b)$ retreives $z$-value from position $b$ of patch $a$ and $f$ is an indicator function that checks if the depth difference between two patches is larger than a threshold $\tau$. 

The original LINEMOD is not scale invariant, which results in a large amount of templates sampled from scale space. Inspired by~\cite{shotton2013real}, we achieve scale invariance by normalizing feature offsets by its depth value.%
\begin{equation}
\label{eq:depth_sim}
\resizebox{1\hsize}{!}{$
\begin{cases}
	S(\mathcal{X}, \mathbb{T}) &= \sum \limits_{ r \in \mathcal{P}} f(\mathcal{X}, \mathcal{O}, c, r)  g( \ori(\mathcal{X}, \frac{r}{D(\mathcal{X},c)}), \ori(\mathcal{O}, \frac{r}{D(\mathcal{O},c)}) ), \\
	f(\mathcal{X}, \mathcal{O}, c, r) &= \delta ( |(D(\mathcal{X}, c) - D(\mathcal{X}, \frac{r}{D(\mathcal{X},c)})) - (D(\mathcal{O}, c) - D(\mathcal{O}, \frac{r}{D(\mathcal{O},c)}))| < \tau )
\end{cases}
$}
\end{equation}
During training, at each split node, we randomly choose a patch as a template $\mathbb{T}$ and measure its similarity with all other patches. Those patches with similarity larger than a threshold $\tau$ will go to one child node, whilst the rest  will go to the other. This split is measured with information gain and repeated multiple times until an optimal one is found.

\subsubsection{Constructing Leaf Nodes}
The training data is recursively split by this process until the tree has reached a maximum depth or the number of samples arriving at a node fall below a threshold. When either one of these criteria is met a leaf node is formed from the patches reaching it. As to the information gain, since no negative information is available during training, thus, we cannot use a classification term but only the regression one for measuring information gain. However, differing from a regression forest, \textit{Latent-Class Hough Forests} still store the class distribution in leaf nodes. Following the approach of Girshick \etal \cite{girshick2011efficient} we only store the indexes of the training patches that reached at the each leaf node and the modes of the distribution, which we efficiently calculate via the MeanShift algorithm. We create a class distribution at the leaf, however, as no background information reaches the leaves during training this distribution is initialized to $p_{fg}^{l} = 1$ and $p_{bg}^{l} = 0$ for the foreground and background probabilities, respectively.
\subsection{Inference}
\label{sec:inference} 
After learning, we have a \textit{Latent-Class Hough Forest} trained with positive samples only. In inference stage, we propose an iterative algorithm to datamine the negative information.
\begin{figure*}[ht]
	\begin{center}
	\includegraphics[width=0.65\linewidth]{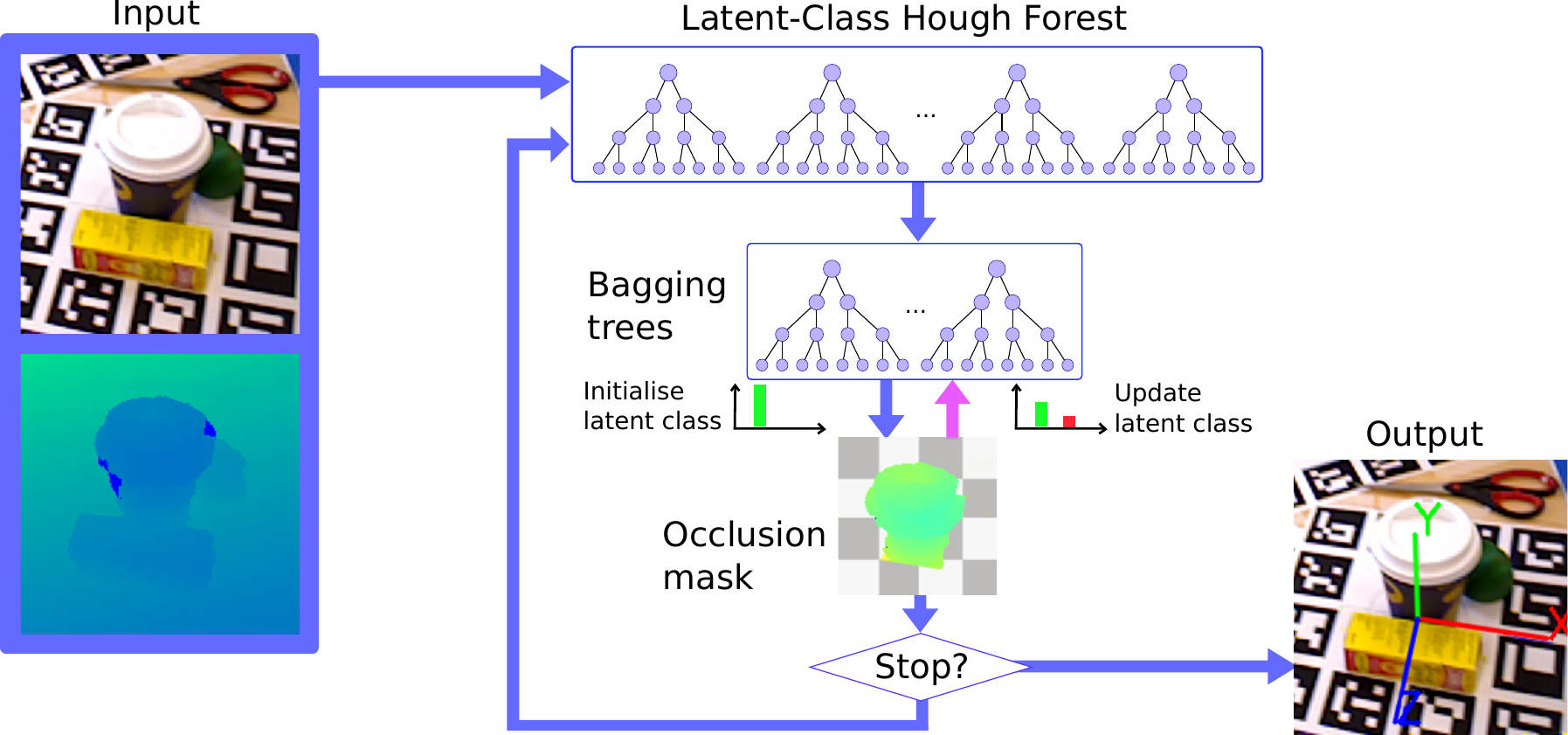}
	\caption{Inference process: Input LINEMOD patches are extracted from RGB-D images. For each iteration, a subset of trees are drawn from the forest with bagging. }
	\label{fig:inference}
	\end{center}
\end{figure*}

\subsubsection{Hough Voting}
Let $E(\theta)$ represent the probability of the random event that the target object exists in the scene under the 6D pose $\theta=(\theta_x, \theta_y, \theta_z, \theta_{roll}, \theta_{pitch}, \theta_{yaw})$. The goal of the voting process is to aggregate the conditional probabilities $p(E(\theta)|\mathcal{P})$ for each patch $\mathcal{P}$, given that each vote originates from a foreground patch ($p_{fg}=1$). For a patch $P$ reaching leaf node $l$ of tree $T$, the conditional probability is formalized as follows:

\begin{equation}
\label{eq:hough_vote}
\begin{split}
p\left(E\left(\theta\right) \vert \mathcal{P} \text{; }\mathcal{T}\right) &= p\left(E\left(\theta\right), p_{fg}^l = 1 \vert \mathcal{P}\right)\\
&= p\left(E\left(\theta\right) \vert p_{fg}^l = 1, \mathcal{P}\right) \cdot p\left(p_{fg}^l = 1 | \mathcal{P}\right)
\end{split}
\end{equation}where $p_{fg}^l$ is the foreground probability at the leaf node, $l$. Finally, for a forest, $\mathcal{F}$, we simply average the probabilities over all trees:
\begin{equation}
\label{eq:forest_hough_vote}
p\left(E\left(\theta\right) \vert \mathcal{P} \text{; }\mathcal{F}\right) = \frac{1}{\vert\mathcal{F}\vert} \sum\limits_{t}^{\vert\mathcal{F}\vert}{p\left(E\left(\theta\right) \vert \mathcal{P} \text{; }\mathcal{T}_t\right)}
\end{equation}
The first factor, $p\left(E\left(\theta\right) \vert p_{fg} = 1, \mathcal{P}\right)$, can be estimated by passing each patch down the forest and accumulating the votes stored at the leaf, in which votes from multiple trees can be combined in an additive manner, this gives us the same probabilities as in Eq. \ref{eq:forest_hough_vote} up to a constant factor. The estimation is then deferred to the ability of locating local maxima in this aggregated space. Traditionally there are two different methods of locating targets in vote space. One is to aggregate all votes in the same position of vote space, and return the peaks with non-maximum suppression~\cite{gall2011hough}. The other is to treat each vote as a data point, and then use MeanShift to locate the mode. In this case an assumption of only one instance in the scene is made to avoid local minimum~\cite{fanelli2011real}. The former is efficient for locating but less accurate than the latter, especially when the votes are sparsely distributed.

To accommodate both efficiency and accuracy, we propose a 3-stage localization technique in which we first aggregate all votes in 2D image space, use this as a score function and locate the top $N$ vote peaks as valid hypotheses; and then use them as initialization for MeanShift to locate modes in 3D translation, $(\theta_x, \theta_y, \theta_z)$; finally find the mode in rotation,$(\theta_{roll}, \theta_{pitch}, \theta_{yaw})$ given translation.

\subsubsection{Update Class Distribution}
 As mentioned in Section~\ref{sec:learning}, class distributions are initialized as all foreground. For each pixel $\mathbf{x}=(\mathcal{P},l)$, located at the center position $l$ of patch $\mathcal{P}$, the initial probability of being foreground is $p_{fg}^\mathbf{x}= 1$. Inspired by~\cite{leibe2004combined}, for each valid hypothesis $\theta$, we backproject to obtain a consensus patch set, i.e. peak of the probability distribution in the Hough space where multiple patches cast their vote, is considered as valid only in cases where its score is above a threshold. This threshold value is different for each object while, our three stage localization technique includes also thresholds per stage. All the thresholds are object-specific and are estimated via trial and error. All consensus voting patches are considered to be foreground and the rest background. 
\begin{equation}
	\label{eq:backproject}
	p_{fg}^\mathbf{x} = \delta ((\sum \limits_{\theta} (p_{fg}^\mathbf{x}|\theta)p(\theta))>0),
	\end{equation}
in which $p(\theta)$ indicates us whether this hypothesis is valid or not and $(p_{fg}^\mathbf{x}|\theta)$ suggests whether $\mathbf{x}$ has voted for $\theta$. As long as $\mathbf{x}$ has voted for one valid hypothesis, it is, then, considered as a foreground pixel ($\delta(\cdot)=1$). In other words, after a valid hypothesis is found via patch voting, we move down to pixel level, where patches that voted for this hypothesis are back-projected to the scene. Since we have an estimation of the object's center in the scene and we know also its diameter, pixels (of the voting patches) that are not spatial consistent (distance to the object centre larger than the diameter) are considered to belong to the background and the rest to the foreground. With this we can update the latent class distribution of each leaf node $l$ by:
\begin{equation}
	\label{eq:update}
	p_{fg}^\mathbf{l} = \frac{\sum \limits_\mathbf{x} p_{fg}^\mathbf{x}}{|\mathbf{x}|}, \mathbf{x} \in l,
\end{equation}
which can be translated to the calculation of the normalized portion of foreground patches that arrives at this leaf node.

\subsubsection{Iterative Process}
The estimation obtained with previous step is inaccurate and uncertain. However, with an iterative process, we update the probability distribution $p_{fg}^\mathbf{l}$ for every iteration. To avoid error accumulation of each round, we propose to draw a subset of trees with bagging, such that not all the trees are updated by previous iteration. The whole process is illustrated in Figure~\ref{fig:inference} and described in Algorithm~\ref{alg:inference}.

\subsubsection{Final segmentation}
With the estimated hypotheses $\{\theta\}$ and a foreground probability mask $\mathcal{Z}$ generated with $p_{fg}^\mathbf{x}$, we can easily obtain a final occlusion-aware segmentation by
\begin{equation}
	\mathcal{M} = \mathbb{B}(\theta)) \cap \mathcal{Z},
\end{equation}
where $\mathbb{B}$ is a bounding box centered at $\theta$. This is helpful for further refinement step such as ICP, which does not work well when occlusion presents.


\begin{algorithm}
\caption{Inference process}
\label{alg:inference}
\begin{algorithmic}[1]
\Require{An input image $\mathcal{I}$; A Hough Forest $\mathcal{F}$}
	\Repeat
	\State Draw a subset of trees $\mathcal{F}^*$ with bagging
	\State Randomly sample a set of patches $P$ from $I$
	\State Propagate $P$ down $\mathcal{F}^*$ and vote for $\theta$ (Eq.~\ref{eq:forest_hough_vote})
	\State Backproject to obtain a foreground mask (Eq.~\ref{eq:backproject})
	\State Partition $P$ into positive and negative subsets by foreground mask
		\[
			P^{fg} = \{\mathbf{x}|\mathbf{x}\in Z\}
		\]
		\[
			P^{bg} = P \backslash P^{fg}
		\]
	\State Update the probabilities at leaf nodes (Eq.~\ref{eq:update}) given $P^{fg}$ and $P^{bg}$.
	\Until{Maximum iteration}
\end{algorithmic}
\end{algorithm}

\section{Experiments}
\label{sec:experiments}
We perform experiments on three 3D pose estimation datasets. The first is the publicly available dataset of of Hinterstoisser \etal ~\cite{hinterstoisser2013model}, which contains 13 distinct objects each associated with an individual test sequence comprising of over 1,100 images with close and far range 2D and 3D clutter. Each test image is annotated with ground truth position and 3D pose. We also introduce our two new datasets, called \textit{Domestic Environments Dataset} and \textit{Bin-picking Dataset} to further evaluate the efficiency of our method in real scenarios. In all tests we use the metric defined in \cite{hinterstoisser2013model} to determine if an estimation is correct. More formally, for a 3D model $\mathcal{M}$, with ground truth rotation $\boldsymbol{R}=(\theta_{pitch}, \theta_{yaw}, \theta_{roll})$ and translation $\boldsymbol{T}=(\theta_{x}, \theta_{y}, \theta_{z})$, given an estimated rotation, $\boldsymbol{\hat{R}}=(\hat{\theta}_{pitch}, \hat{\theta}_{yaw}, \hat{\theta}_{roll})$ and translation, $\boldsymbol{\hat{T}}=(\hat{\theta}_{x}, \hat{\theta}_{y}, \hat{\theta}_{z})$, the matching score is defined as

\begin{equation}\label{non_sym_eq}
m = \underset{\boldsymbol{x} \in \mathcal{M}}\avg ||(\boldsymbol{R}\boldsymbol{x} + \boldsymbol{T}) - (\boldsymbol{\hat{R}}\boldsymbol{x} + \boldsymbol{\hat{T}})||
\end{equation}for non-symmetric objects and
\begin{equation}\label{sym_eq}
m = \underset{{\boldsymbol{x_1} \in \mathcal{M}}}\avg \underset{\boldsymbol{x_2} \in \mathcal{M}}\min||(\boldsymbol{R}\boldsymbol{x_1} + \boldsymbol{T}) - (\boldsymbol{\hat{R}}\boldsymbol{x_2} + \boldsymbol{\hat{T}})||
\end{equation}for symmetric objects. An estimation is deemed correct if $m \leq k_m*d$, where $k_m$ is a chosen coefficient and $d$ is the diameter of $\mathcal{M}$. We would like to note that the symmetric version (Eq. \ref{sym_eq}) of the matching score assumes complete symmetry around an axis, like bowls. However, other objects that  with, for example, square shape have an advantage if assumed symmetric.  
 
Unlike \cite{hinterstoisser2013model}, in which only the top N detections from each image are selected, we also compute precision-recall curves and present the F1-Score which is the harmonic mean of precision and recall. We argue that this is a more accurate form of comparison, as directly comparing detections is inaccurate as some images may be more challenging than others and the number of target objects may be unknown (as is the case in our new datasets).

In the proposed method a forest is currently trained for one object while there are several works (e.g. \cite{brachmann2014learning,doumanoglou20156d}) that utilize a single forest for multiple objects. Our method could be extended to multiple objects per forest by adding three different objective functions that would be used for measuring information gain. During training, similar to classic Hough Forests~\cite{gall2011hough}, we could randomly select to perform either entropy minimization of the class distribution or entropy minimization of the of the  parameters. Practically this means that we could perform multi-class classification and 6D object pose estimation, addressing, thus, one shortcoming of our method that assumes at least one object instance to always be present in the scene.   

In Sec. \ref{sec:fram_param} we present extensive experiments on the various parameters of the framework such as as patch size, number of trees and number of iterations to infer class distributions at test time. In Sec. \ref{sec:comp_self} we perform self comparison tests highlighting the benefits of adding scale-invariance to the template similarity measure (Eq. \ref{eq:scale_inv_similarity}) and using co-training to update the latent class distributions (Algorithm \ref{alg:inference}). Following this, in Sec. \ref{sec:comp_LINEMOD} we compare the performance of our method on the famous dataset of Hinterstoisser \etal ~\cite{hinterstoisser2013model} against the state of the art works of LINEMOD \cite{hinterstoisser2011multimodal}, Drost \etal ~\cite{drost2010model}, Rios Cabrera \etal ~\cite{Rios_Cabrera_2013_ICCV} and Brachmann \etal ~\cite{brachmann2014learning}. Moreover, in Sec. \ref{sec:comp_Domestic} and \ref{sec:comp_Bin_Picking} we present our new datasets and additional comparative analysis. 

\subsection{Framework Parameters}
\label{sec:fram_param}

Parameter optimization was performed on a validation dataset that was created by randomly selecting a subset of our own datasets. For each object class, we train a \textit{Latent-Class Hough Forest} with varying number of trees and patch sizes. Moreover, during inference and for the co-training stage we experiment with different number of iterations, while the number of hypothesis to be backprojected per iteration is set $N = 10$. We choose 10 as it is greater than the number of instances present in all datasets, however this number is not fixed and can be adapted based on the application. Furthermore, in all experiments for parameter optimization the coefficient $k_m$ is set to the value of 0.15, the results with this coefficient are also found to be visually correct.

\begin{figure}[h]
\begin{center}
\includegraphics[width=1\linewidth]{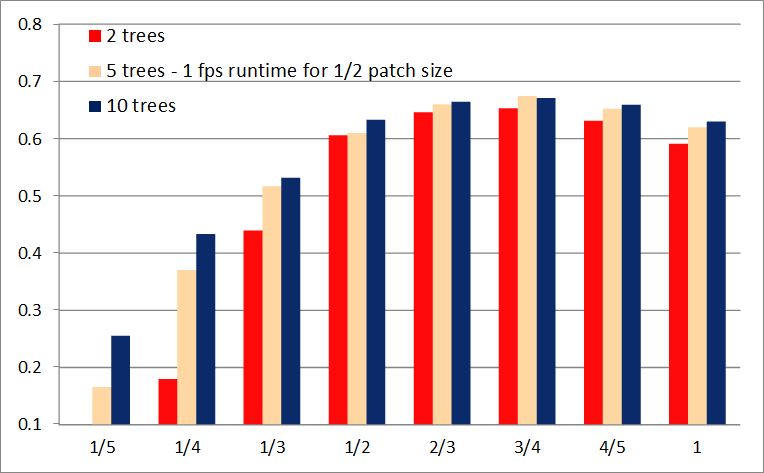}
\end{center}
\caption{F1-Scores for the 6 different patch sizes and 3 different forests in our validation dataset.}\label{fig:patch_analysis}
\end{figure}

\subsubsection{Patch Size and Number of Trees}
Selecting the size of the patch is of paramount importance since large patches tend to match the disadvantages of a holistic template (i.e. sensitive clutter, occlusions etc.) while small ones are prone to noise. The size of the patch depends on the object's identity and it is relative to its size. We extract the bounding box of the target object by rendering its 3D mesh model and modify the patch size to be proportional to the size of the bounding box. Experimenting with 6 different patch sizes, as shown in Fig. \ref{fig:patch_analysis}, revealed that increasing the size of the patch has a direct effect on the F1-Score, while saturation occurs for patches with size equal to 2/3 of the bounding box and higher. Patches resembling holistic ones (i.e. 4/5 and 1) are proven to be very prone to occlusions. 


We have additionally experimented with the number of the trees of our \textit{Latent-Class Hough Forests}. As seen in Fig. \ref{fig:patch_analysis}, given a patch size of 2/3 and larger, then selecting more than ten trees is meaningless. It is apparent that, using patches with relatively large size and forests with more than five trees, puts additional computational burden to our system. For instance, we can achieve real-time execution (1 fps) if we use 1/2 as patch size, 2 trees and zero iterations at the co-training stage. However, selecting larger patches and forests results in a respective drop of the execution time. 

Regarding the balance of the trees, the learned thresholds $\tau$ expand through the whole range of the values of the split function in a certain node. Usually, the nodes at the top of the tree tend to be more balanced, since there is a variety of different patches, whereas nodes at lower depths are less balanced. However, as the tree grows, each node contains less patches and balancing depends on how well the available patches can be split in two sets, according to our objective function. In fact, producing balanced nodes is not always the best way to split the samples, since it depends on the feature space the patches lay on. The forest optimizes splitting according to the semantic appearance of the patches, and balanced splitting does not always satisfy this objective. However, we have noticed that for objects “Coffee Cup”, “Juice Carton” and “Joystick” the percentage of training patches that move to the child with the lowest number of samples is around 43\%, implying well balanced trees, while for the rest of the objects this percentage is around 35\%. Another way to measure the balance of the trees was proposed in \cite{ristin2014incremental} where instead of taking one patch, one could also take two or more patches and assign the patch to closest one. 

\begin{figure}[h]
\begin{center}
\includegraphics[width=1\linewidth]{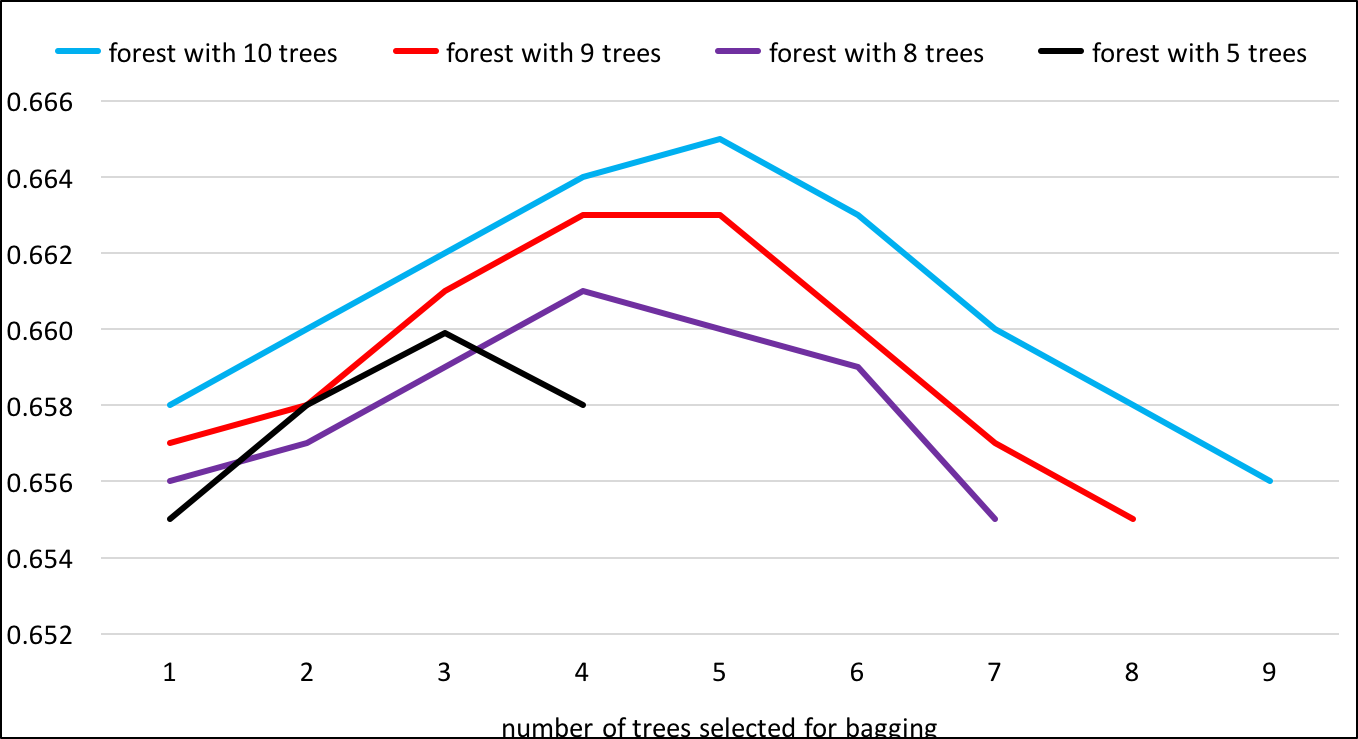}
\end{center}
\caption{The impact of bagging in our iterative process. Accuracy (F1-Scores) vs. number of trees selected for bagging under four different forest sizes.}\label{fig:bagging_analysis}
\end{figure}

\begin{figure}[h]
\begin{center}
\includegraphics[width=1\linewidth]{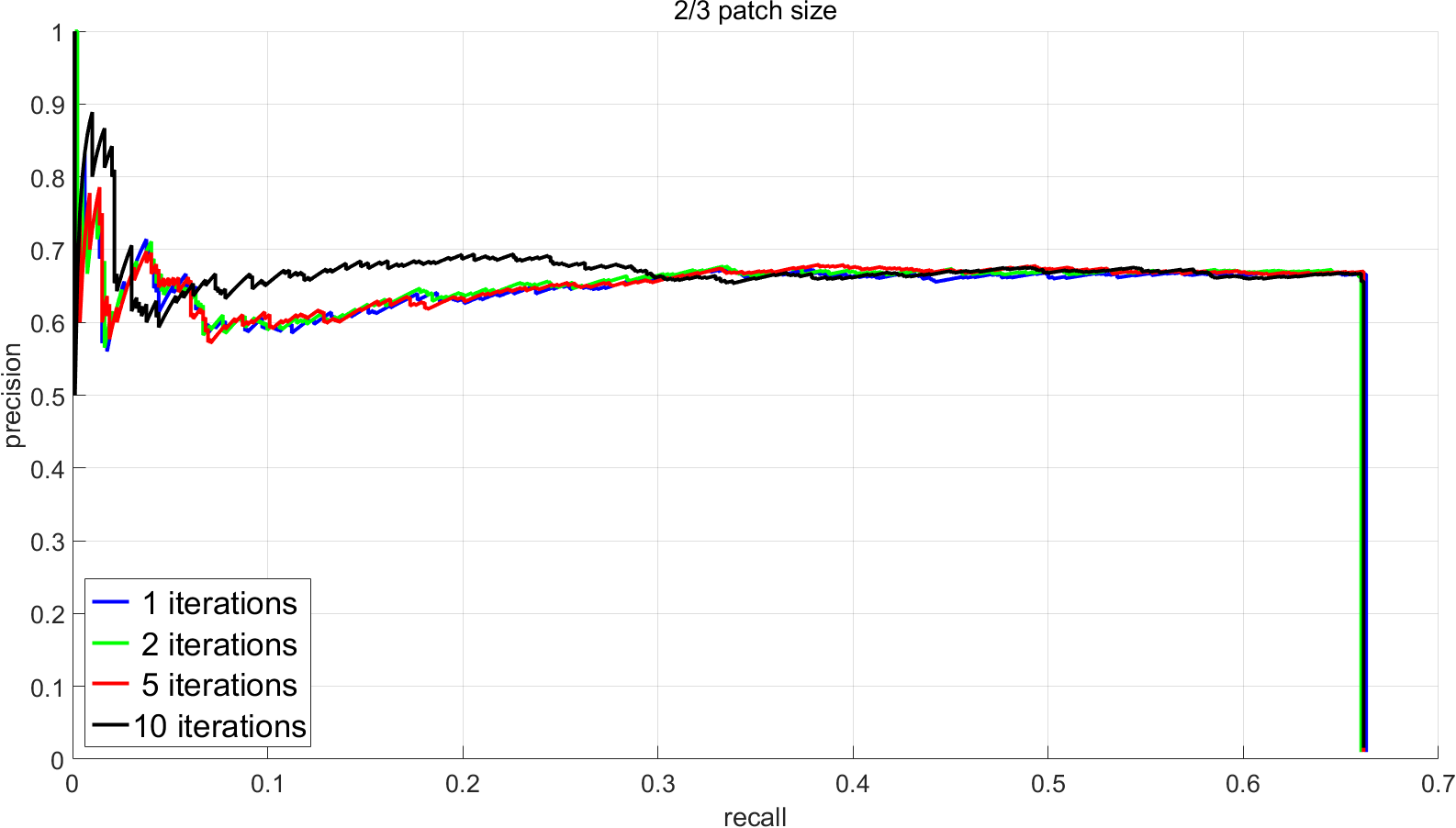}
\end{center}
\caption{The impact of the number of iterations on the performance of our system for our validation dataset on a fixed patch size of 2/3 and a forest with 10 trees.}\label{fig:tren_iter}
\end{figure}

\begin{figure*}[ht]
\begin{center}
\subfigure{\includegraphics[width=0.495\linewidth]{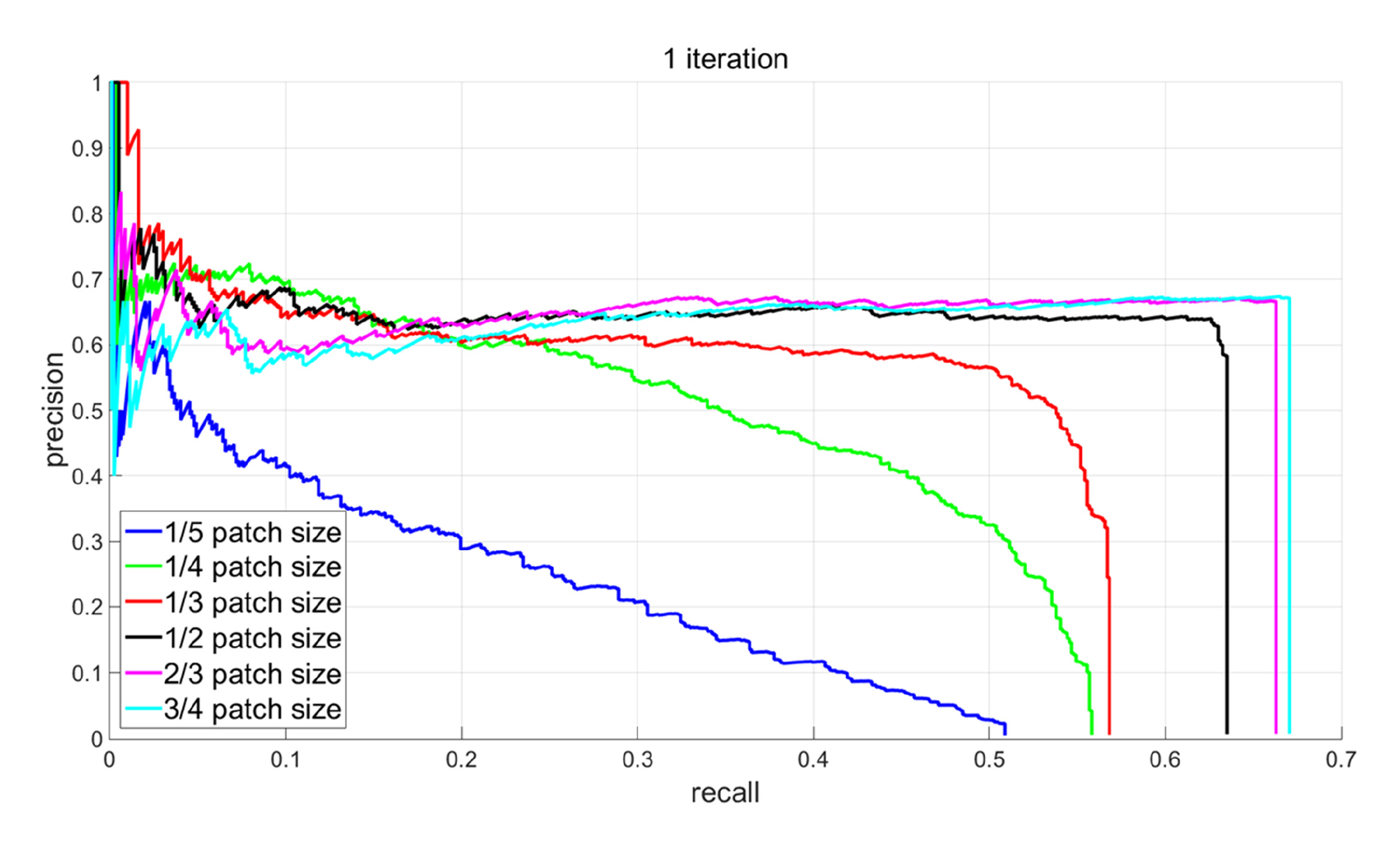}}
\subfigure{\includegraphics[width=0.495\linewidth]{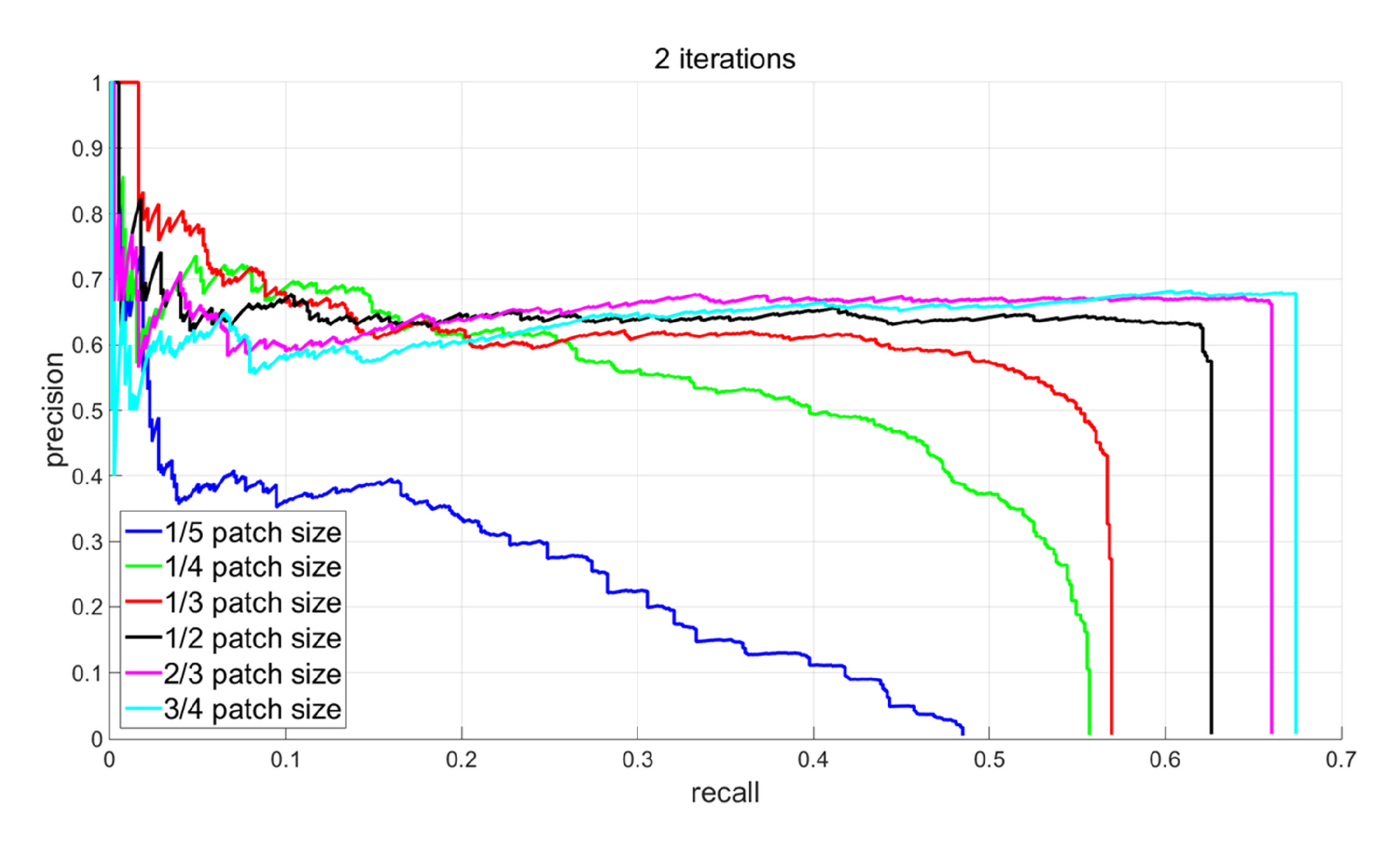}}
\subfigure{\includegraphics[width=0.495\linewidth]{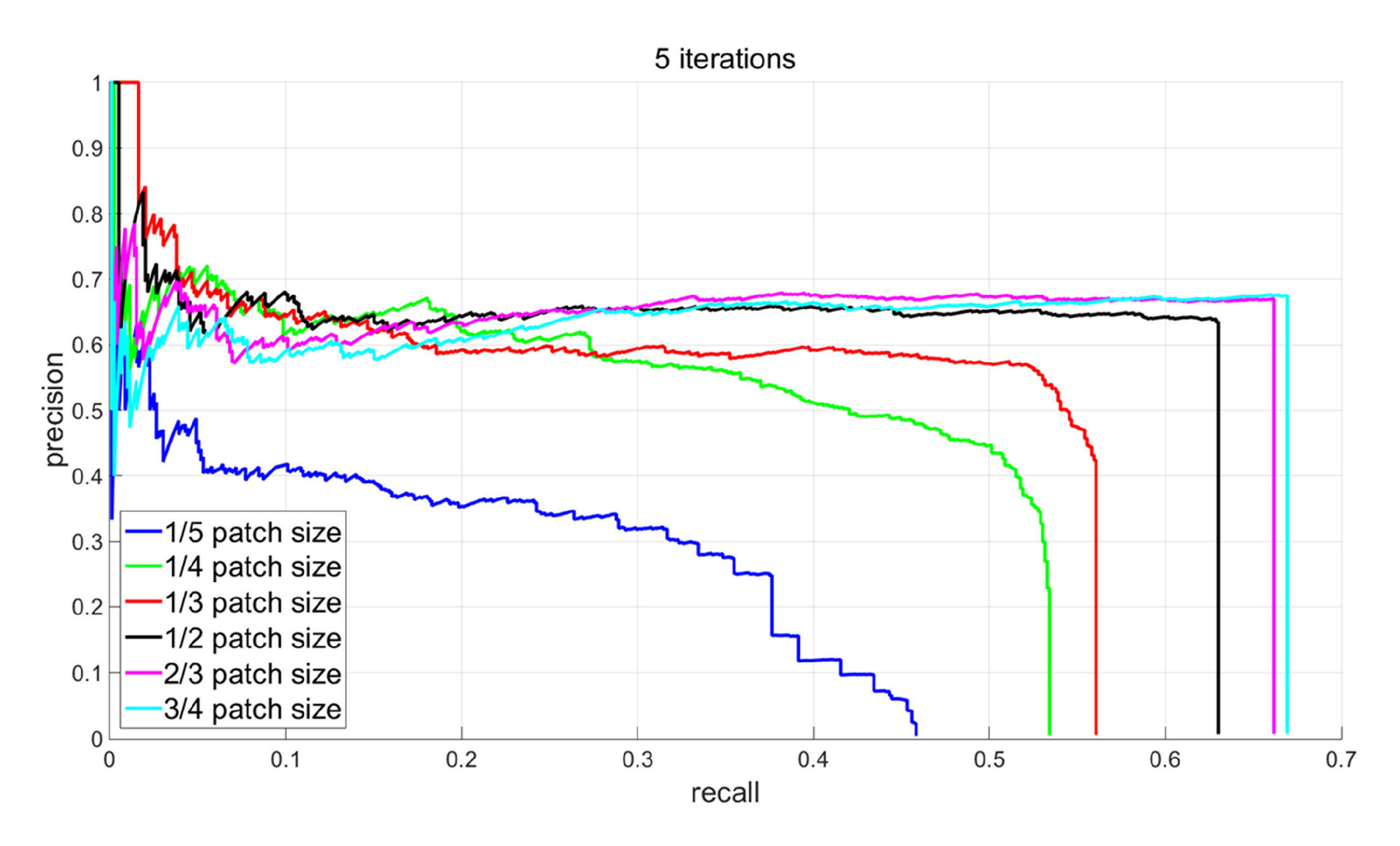}}
\subfigure{\includegraphics[width=0.495\linewidth]{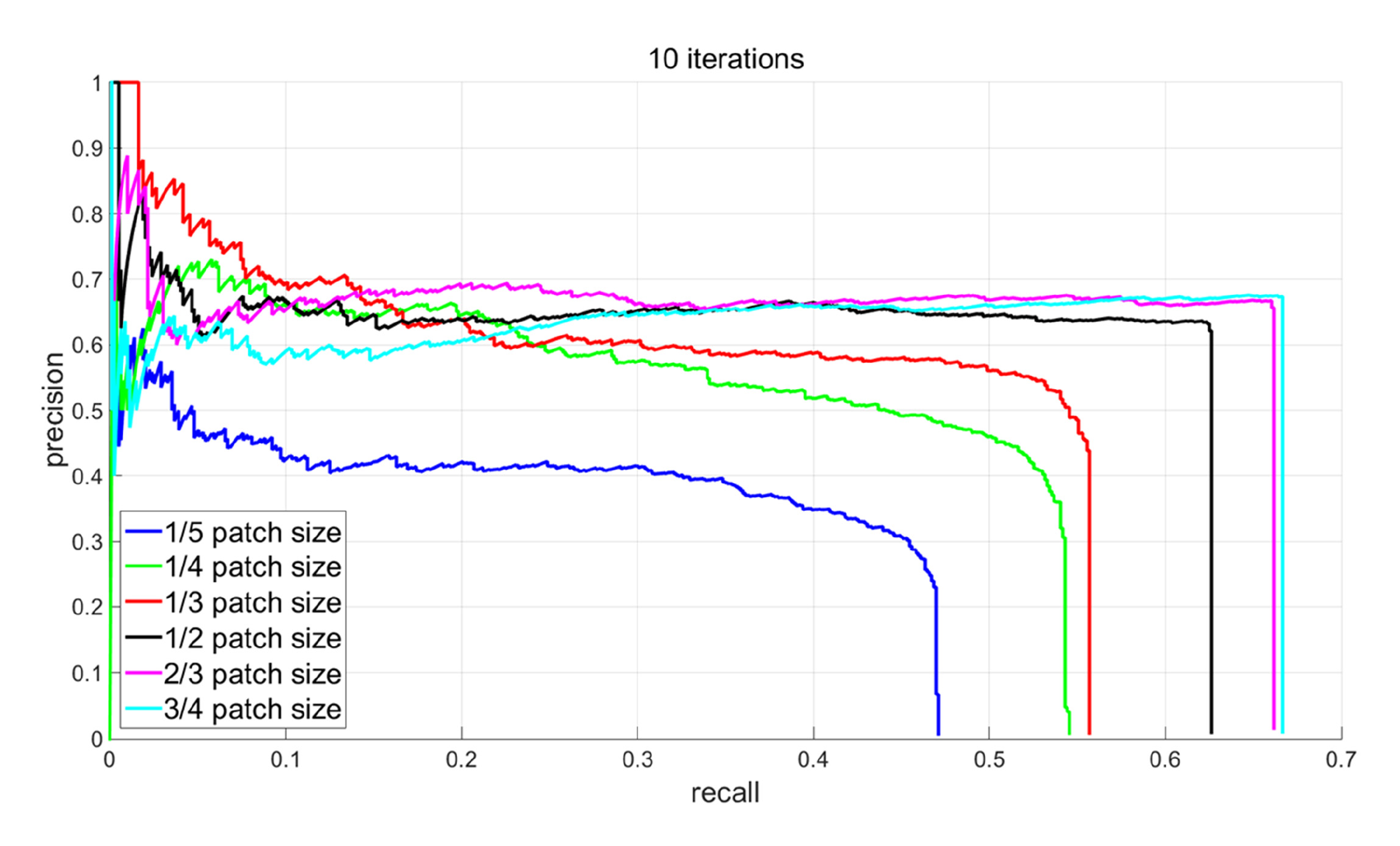}}
\end{center}
\caption{Precision-Recall curves for different number of iterations and patch sizes.}\label{fig:iter_patch}
\end{figure*}

\subsubsection{Bagging and Number of Iterations}
During inference we randomly draw a subset of trees with bagging and iteratively update class distributions in order to datamine the negative information. The impact of bagging is shown in Fig. \ref{fig:bagging_analysis}. Given four different sizes of forests (i.e. 10, 9, 8 and 5 trees) we randomly select a subset of trees to obtain an initial object hypotheses set that is then used to construct a consensus pixel test. To investigate the effect of the number of selected trees, we compared the performance levels of our method when drawing a) 1 to 9 trees from a 10 tree forest, b) 1 to 8 trees from a 9 tree forest, c) 1 to 7 trees from an 8 tree forest and d) 1 to 4 from a 5 tree forest. According to the conducted experiments (see Fig. \ref{fig:bagging_analysis}), our method preforms better in cases we split the forest in the middle, thus, creating two classifiers of equal size.    

The quality of the resulted segmentation mask depends on the number of iterations of the co-training process. Since our method makes no use of negative data during training, one straightforward question is whether training with both positive and negative patches would reduce the number of iterations required for extracting the segmentation mask. Towards this end, we used the background dataset of~\cite{brachmann2014learning} that contains RGB-D images of different cluttered office backgrounds. Patches were sampled from the negative data, pushed through the trees and the classes’ probability distributions were updated. As expected the overall accuracy was not improved while a slightly faster convergence was noticed only in very few cases. After the first iteration, the extracted consensus pixel set contains both positive and negative data that come directly from the observable unlabelled data in the scene. We believe that this is the reason why training with negative samples offers slightly faster convergence only in very few cases. Fig. \ref{fig:tren_iter} illustrates the effect of the iterative part on the performance of our method on our validation dataset. As expected, the iterative method affects only the precision of the system and not the recall, since it does not serve as a false positive rejection framework. Higher number of iterations offer slightly better accuracy with the cost, however, of additional execution time.

In turn, Fig. \ref{fig:iter_patch} illustrates the conditional effect of different patch sizes for different number of iterations during inference. At this point we would like to note that, as seen in Table \ref{tab:comp_to_state_of_art_hinter_f1_scores}, our system is capable of producing accurate 6 DoF pose estimations even without the iterative part. However, depending on the application domain we might switch to different parameters to meet the respective demands. For example, for industrial object manipulation, real-time execution usually constitutes a prerequisite, while robots operating in domestic environments, emphasize more on the accuracy levels. Towards this end, for the \textit{Domestic Environments Dataset} we set the patch size of 2/3, use a forest of 10 trees and set the number of iterations to 10, while for our \textit{Bin-picking Dataset}, we investigate both the aforementioned parameters and the effect of smaller patches (1/2) for forests with only 2 trees without any iteration at all. 

Our approach is not guaranteed to converge and it is possible for our method to produce a solution where all the foreground probabilities converge to zero. However, we have not noticed such a case throughout the exhaustive experimental analysis. The extracted consensus pixel set steams from patches that voted for a hypothesis with a score larger than a threshold, which is different for each object and estimated via trial and error. However, false positives play an important role in our method and can generate false segmentation masks, as shown in the last row of Fig. \ref{fig:qual_res}.

\begin{table}[h]
\small
\centering
\caption{Impact of different modalities on the performance of our method based on a smaller validation subset.}
\label{tab:comp_modal}
\begin{tabular}{lc|c|c|c}
\multicolumn{1}{l|}{\textbf{Object}}	&\multicolumn{1}{l|}{RGB}		&\multicolumn{1}{l|}{Depth}	&\multicolumn{1}{l|}{RGB-D} \\ \hline
\multicolumn{1}{l|}{Coffee Cup}		&0.748					&\textbf{0.838}				&0.828		\\ 
\multicolumn{1}{l|}{Camera}		&0.309					&0.324					&\textbf{0.369}		\\      
\end{tabular}
\end{table}

\begin{table*}[!htb]
\begin{center}
\caption{Matching score and speed on the dataset of Hinterstoisser \etal ~\cite{hinterstoisser2013model} for LINEMOD \cite{hinterstoisser2011multimodal}, the methods of Drost \etal ~\cite{drost2010model}, Rios Cabrera \etal ~\cite{Rios_Cabrera_2013_ICCV}, Brachman \etal ~\cite{brachmann2014learning} and our approach. In the second column we report which matching score (Eq. \ref{non_sym_eq} or Eq. \ref{sym_eq}) was used.}\label{tab:comp_to_state_of_art_hinter_percentages}
\begin{tabular}{|l|c|c|c|c|c|c|}
\hline
\multicolumn{2}{|c|}{\textbf{Approach}}           & \textbf{LINEMOD \cite{hinterstoisser2011multimodal}} & \textbf{Drost \etal ~\cite{drost2010model}} & \textbf{Rios Cabrera \etal ~\cite{Rios_Cabrera_2013_ICCV}} & \textbf{Brachmann \etal ~\cite{brachmann2014learning}} & \textbf{Our Approach} \\ \hline
\textbf{Sequence (\# images)} & \textbf{Metric} & \multicolumn{5}{|c|}{\textbf{Matching Score / Speed (when available)}} \\ \hline
Ape(1235)               	&\ref{non_sym_eq}		&\multicolumn{1}{|c|}{95.8\% / 127ms}	&\multicolumn{1}{|c|}{86.5\% / 22.7s}		&\multicolumn{1}{|c|}{95.0\% / 55.8ms}		&\multicolumn{1}{|c|}{\textbf{95.8\% / -}}	&\multicolumn{1}{|c|}{95.7\% / 1.82s}			\\ \hline
Bench Vise (1214)    &\ref{non_sym_eq}		&\multicolumn{1}{|c|}{98.7\% / 115ms}	&\multicolumn{1}{|c|}{70.7\% / 2.94s}		&\multicolumn{1}{|c|}{98.9\% / 53.3ms}		&\multicolumn{1}{|c|}{\textbf{100\% / -}}	&\multicolumn{1}{|c|}{99.7\% / 2.11s}			\\ \hline
Driller (1187)       	&\ref{non_sym_eq}		&\multicolumn{1}{|c|}{93.6\% / 121ms}	&\multicolumn{1}{|c|}{87.3\% / 2.65s}		&\multicolumn{1}{|c|}{94.3\% / 54.6ms}		&\multicolumn{1}{|c|}{\textbf{99.5\% / -}}	&\multicolumn{1}{|c|}{99.2\% / 1.91s}			\\ \hline
Cam (1200)        	&\ref{non_sym_eq}		&\multicolumn{1}{|c|}{97.5\% / 148ms}	&\multicolumn{1}{|c|}{78.6\% / 2.81s}		&\multicolumn{1}{|c|}{98.2\% / 58.4ms}		&\multicolumn{1}{|c|}{\textbf{99.6\% / -}}	&\multicolumn{1}{|c|}{\textbf{99.6\% / 1.98s}}	\\ \hline
Can (1195)         	&\ref{non_sym_eq}		&\multicolumn{1}{|c|}{95.4\% / 122ms}	&\multicolumn{1}{|c|}{80.2\% / 1.60s}		&\multicolumn{1}{|c|}{\textbf{96.3\% / 55.3ms}}	&\multicolumn{1}{|c|}{95.9\% / -}		&\multicolumn{1}{|c|}{96.1\% / 2.08s}			\\ \hline
Iron (1151)         	&\ref{non_sym_eq}		&\multicolumn{1}{|c|}{97.5\% / 116ms}	&\multicolumn{1}{|c|}{84.9\% / 3.18s}		&\multicolumn{1}{|c|}{98.4\% / 54.3ms}		&\multicolumn{1}{|c|}{97.6\% / -}		&\multicolumn{1}{|c|}{\textbf{98.5\% / 1.97s}}	\\ \hline
Lamp (1226)     	&\ref{non_sym_eq}		&\multicolumn{1}{|c|}{97.7\% / 125ms}	&\multicolumn{1}{|c|}{93.9\% / 2.29s}		&\multicolumn{1}{|c|}{97.9\% / 54.8ms}		&\multicolumn{1}{|c|}{\textbf{99.8\% / -}}	&\multicolumn{1}{|c|}{99.6\% / 2.01s}			\\ \hline
Phone (1224)      	&\ref{non_sym_eq}		&\multicolumn{1}{|c|}{93.3\% / 157ms}	&\multicolumn{1}{|c|}{80.7\% / 4.70s}		&\multicolumn{1}{|c|}{95.3\% / 58.4ms}		&\multicolumn{1}{|c|}{\textbf{97.6\% / -}}	&\multicolumn{1}{|c|}{96.7\% / 2.05s}			\\ \hline
Cat (1178)         	&\ref{non_sym_eq}		&\multicolumn{1}{|c|}{99.3\% / 111ms}	&\multicolumn{1}{|c|}{85.4\% / 7.52s}		&\multicolumn{1}{|c|}{99.1\% / 53.5ms}		&\multicolumn{1}{|c|}{\textbf{100\% / -}}	&\multicolumn{1}{|c|}{99.8\% / 1.97s}			\\ \hline
Hole Punch (1236)  	&\ref{non_sym_eq}		&\multicolumn{1}{|c|}{95.9\% / 110ms}	&\multicolumn{1}{|c|}{77.4\% / 8.30s}		&\multicolumn{1}{|c|}{97.5\% / 54.2ms}		&\multicolumn{1}{|c|}{\textbf{99.4\% / -}}	&\multicolumn{1}{|c|}{99.5\% / 1.92s}			\\ \hline
Duck (1253)     		&\ref{non_sym_eq}		&\multicolumn{1}{|c|}{95.9\% / 104ms}	&\multicolumn{1}{|c|}{40.0\% / 6.97s}		&\multicolumn{1}{|c|}{94.2\% / 53.6ms}		&\multicolumn{1}{|c|}{95.9\% / -}		&\multicolumn{1}{|c|}{\textbf{96.1\% / 1.74s}}	\\ \hline
Box (1252)       		&\ref{sym_eq}	&\multicolumn{1}{|c|}{99.8\% / 101ms}	&\multicolumn{1}{|c|}{97.0\% / 2.94s}		&\multicolumn{1}{|c|}{99.8\% / 56.0ms}		&\multicolumn{1}{|c|}{98.0\% / -}		&\multicolumn{1}{|c|}{\textbf{98.1\% / 2.10s}}	\\ \hline
Glue (1219)      		&\ref{sym_eq}	&\multicolumn{1}{|c|}{91.8\% / 135ms}	&\multicolumn{1}{|c|}{57.2\% / 4.03s}		&\multicolumn{1}{|c|}{96.3\% / 58.5ms}		&\multicolumn{1}{|c|}{\textbf{98.9\% / -}}	&\multicolumn{1}{|c|}{98.2\% / 1.83s}			\\ \hline
\textbf{Average}	&		&\multicolumn{1}{|c|}{96.3\% / 122ms}	&\multicolumn{1}{|c|}{78.4\% / 5.58s}		&\multicolumn{1}{|c|}{97.1\% / 55.4ms}		&\multicolumn{1}{|c|}{\textbf{98.3\% / -}}	&\multicolumn{1}{|c|}{98.2\% / 1.96s}			\\ \hline
\end{tabular}
\end{center}
\end{table*}

\begin{table*}[!htb]
\begin{center}
\caption{F1-Scores for LINEMOD \cite{hinterstoisser2011multimodal}, the method of Drost \etal ~\cite{drost2010model} and our approach for each object class for the dataset of Hinterstoisser \etal ~\cite{hinterstoisser2013model} and our \textit{Domestic Environment Dataset}.}\label{tab:comp_to_state_of_art_hinter_f1_scores}
\begin{tabular}{|l|c|c|c|c|c|}
\hline
\textbf{Approach}             & \textbf{LINEMOD \cite{hinterstoisser2011multimodal}} & \textbf{Drost \etal ~\cite{drost2010model}} & \textbf{SI LINEMOD}	& \textbf{Without Iterations} & \textbf{With Iterations} \\ \hline
\multicolumn{6}{|c|}{\textbf{Dataset of Hinterstoisser \etal ~\cite{hinterstoisser2013model}}} \\ \hline
\textbf{Sequence (\# images)} & \multicolumn{5}{|c|}{\textbf{F1-Score}} \\ \hline
Ape(1235)			&\multicolumn{1}{|c|}{0.533}		&\multicolumn{1}{|c|}{0.628}	&\multicolumn{1}{|c|}{0.631}		&\multicolumn{1}{|c|}{0.799}	&\multicolumn{1}{|c|}{\textbf{0.855}}	\\ \hline
Bench Vise (1214)	&\multicolumn{1}{|c|}{0.846}		&\multicolumn{1}{|c|}{0.237}	&\multicolumn{1}{|c|}{0.869}		&\multicolumn{1}{|c|}{0.941}	&\multicolumn{1}{|c|}{\textbf{0.961}}	\\ \hline
Driller (1187)		&\multicolumn{1}{|c|}{0.691}		&\multicolumn{1}{|c|}{0.597}	&\multicolumn{1}{|c|}{0.744}		&\multicolumn{1}{|c|}{0.899}	&\multicolumn{1}{|c|}{\textbf{0.905}}	\\ \hline
Cam (1200)		&\multicolumn{1}{|c|}{0.640}		&\multicolumn{1}{|c|}{0.513}	&\multicolumn{1}{|c|}{0.711}		&\multicolumn{1}{|c|}{0.636}	&\multicolumn{1}{|c|}{\textbf{0.718}}	\\ \hline
Can (1195)		&\multicolumn{1}{|c|}{0.512}		&\multicolumn{1}{|c|}{0.510}	&\multicolumn{1}{|c|}{0.550}		&\multicolumn{1}{|c|}{0.708}	&\multicolumn{1}{|c|}{\textbf{0.709}}	\\ \hline
Iron (1151)		&\multicolumn{1}{|c|}{0.683}		&\multicolumn{1}{|c|}{0.405}	&\multicolumn{1}{|c|}{0.749}		&\multicolumn{1}{|c|}{0.705} 	&\multicolumn{1}{|c|}{\textbf{0.735}}	\\ \hline
Lamp (1226)		&\multicolumn{1}{|c|}{0.675}		&\multicolumn{1}{|c|}{0.776}	&\multicolumn{1}{|c|}{0.790}		&\multicolumn{1}{|c|}{0.911}	&\multicolumn{1}{|c|}{\textbf{0.921}}	\\ \hline
Phone (1224)		&\multicolumn{1}{|c|}{0.563}		&\multicolumn{1}{|c|}{0.471}	&\multicolumn{1}{|c|}{0.655}		&\multicolumn{1}{|c|}{0.660}	&\multicolumn{1}{|c|}{\textbf{0.728}}	\\ \hline
Cat (1178)			&\multicolumn{1}{|c|}{0.656}		&\multicolumn{1}{|c|}{0.566}	&\multicolumn{1}{|c|}{0.773}		&\multicolumn{1}{|c|}{0.884}	&\multicolumn{1}{|c|}{\textbf{0.888}}	\\ \hline
Hole Punch (1236)	&\multicolumn{1}{|c|}{0.516}		&\multicolumn{1}{|c|}{0.500}	&\multicolumn{1}{|c|}{0.601}		&\multicolumn{1}{|c|}{0.819}	&\multicolumn{1}{|c|}{\textbf{0.875}}	\\ \hline
Duck (1253)		&\multicolumn{1}{|c|}{0.580}		&\multicolumn{1}{|c|}{0.313}	&\multicolumn{1}{|c|}{0.659}		&\multicolumn{1}{|c|}{0.888}	&\multicolumn{1}{|c|}{\textbf{0.907}}	\\ \hline
Box (1252)		&\multicolumn{1}{|c|}{0.860}		&\multicolumn{1}{|c|}{0.826}	&\multicolumn{1}{|c|}{\textbf{0.933}}	&\multicolumn{1}{|c|}{0.736}	&\multicolumn{1}{|c|}{0.740}		\\ \hline
Glue (1219)		&\multicolumn{1}{|c|}{0.438}		&\multicolumn{1}{|c|}{0.382}	&\multicolumn{1}{|c|}{0.462}		&\multicolumn{1}{|c|}{0.643}	&\multicolumn{1}{|c|}{\textbf{0.678}}	\\ \hline
\textbf{Average}	&\multicolumn{1}{|c|}{0.630}		&\multicolumn{1}{|c|}{0.517}	&\multicolumn{1}{|c|}{0.702}		&\multicolumn{1}{|c|}{0.788}	&\multicolumn{1}{|c|}{\textbf{0.817}}	\\ \hline
\multicolumn{6}{|c|}{\textbf{Domestic Environment Dataset}}\\ \hline
\textbf{Sequence (\# images)} & \multicolumn{5}{|c|}{\textbf{F1-Score}}  \\ \hline
Coffee Cup (708)	&\multicolumn{1}{|c|}{0.819}		&\multicolumn{1}{|c|}{0.867}	&\multicolumn{1}{|c|}{0.831}		&\multicolumn{1}{|c|}{0.821}	&\multicolumn{1}{|c|}{\textbf{0.877}}	\\ \hline
Shampoo (1058)	&\multicolumn{1}{|c|}{0.625}		&\multicolumn{1}{|c|}{0.651}	&\multicolumn{1}{|c|}{0.649}		&\multicolumn{1}{|c|}{0.712}	&\multicolumn{1}{|c|}{\textbf{0.759}}	\\ \hline
Joystick (1032)		&\multicolumn{1}{|c|}{0.454}		&\multicolumn{1}{|c|}{0.277}	&\multicolumn{1}{|c|}{0.491}		&\multicolumn{1}{|c|}{0.511}	&\multicolumn{1}{|c|}{0.534}		\\ \hline
Camera (708)		&\multicolumn{1}{|c|}{0.422}		&\multicolumn{1}{|c|}{0.407}	&\multicolumn{1}{|c|}{\textbf{0.498}}	&\multicolumn{1}{|c|}{0.291}	&\multicolumn{1}{|c|}{0.372}		\\ \hline
Juice Carton (859)	&\multicolumn{1}{|c|}{0.494}		&\multicolumn{1}{|c|}{0.604}	&\multicolumn{1}{|c|}{0.506}		&\multicolumn{1}{|c|}{0.812}	&\multicolumn{1}{|c|}{\textbf{0.870}}	\\ \hline
Milk (860)			&\multicolumn{1}{|c|}{0.176}		&\multicolumn{1}{|c|}{0.259}	&\multicolumn{1}{|c|}{0.228}		&\multicolumn{1}{|c|}{0.315}	&\multicolumn{1}{|c|}{0.385}		\\ \hline
\textbf{Average}	&\multicolumn{1}{|c|}{0.498}		&\multicolumn{1}{|c|}{0.511}	&\multicolumn{1}{|c|}{0.533}		&\multicolumn{1}{|c|}{0.582}	&\multicolumn{1}{|c|}{\textbf{0.633}}	\\ \hline
\end{tabular}
\end{center}
\end{table*}

\begin{figure*}[!htb]
\begin{center}
  \subfigure{\includegraphics[width=0.45\linewidth]{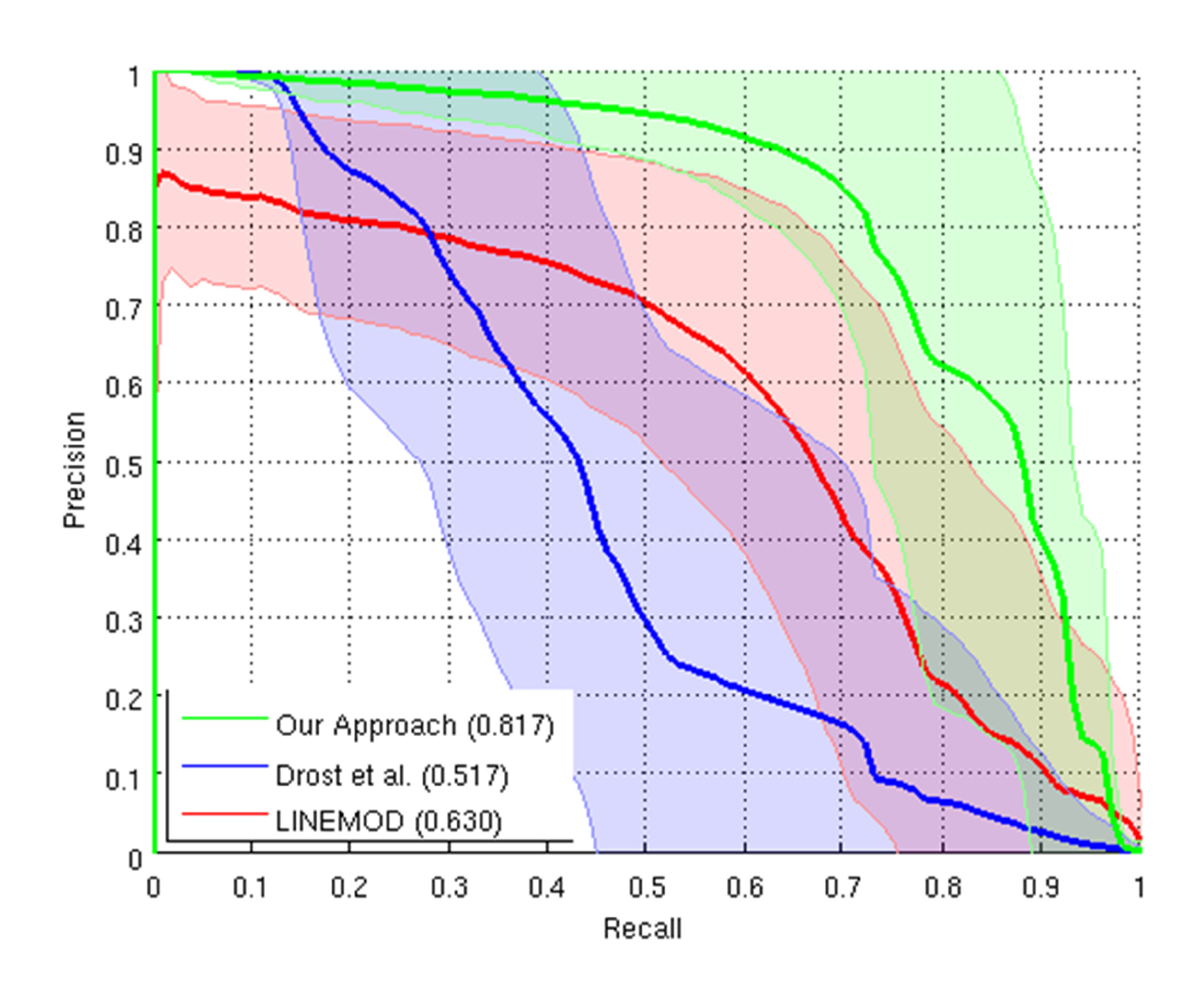}}
    \subfigure{\includegraphics[width=0.45\linewidth]{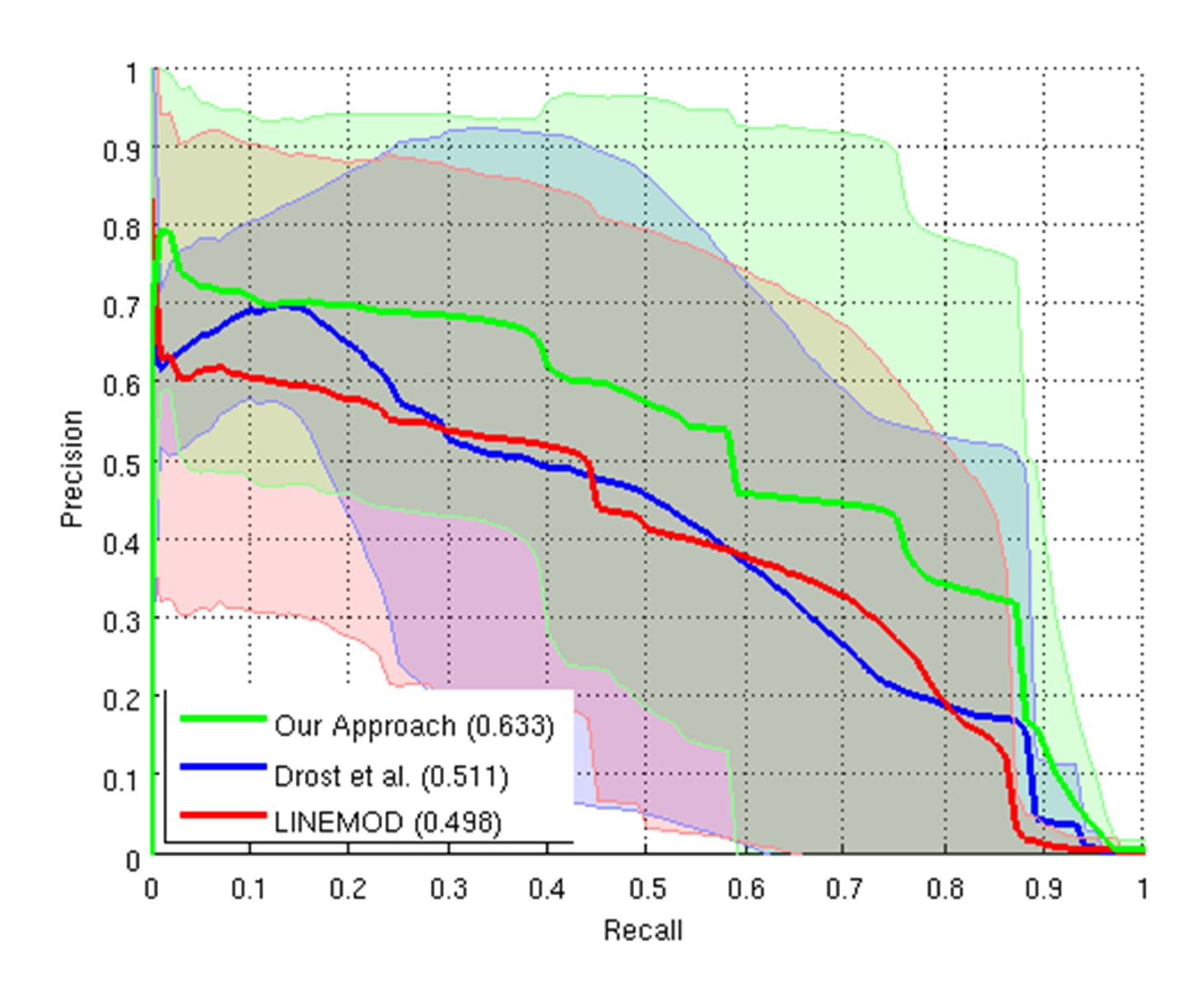}}
    \end{center}
\caption{Average Precision-Recall curve over all objects in the dataset of LINEMOD~\cite{hinterstoisser2013model} (left) and our \textit{Domestic Environment Dataset} (right). The shaded region represents one standard deviation above and below the precision value at a given recall value.}\label{fig:comp_to_state_of_arts}
\end{figure*}

\subsection{Comparative study}\label{sec:comp_state_of_arts}
In the following section we present a comparative analysis of the performance of our method against other state of the art techniques. We perform a self comparisons to investigate the role of our iterative method, while we compare against several recently published methods in three different datasets.

\subsubsection{Self Comparisons}\label{sec:comp_self}
We perform two self comparisons on the dataset of Hinterstoisser \etal ~\cite{hinterstoisser2013model} and our \textit{Domestic Environment Dataset}. Firstly we compare the results of our method with and without the iteratative process. As can be seen in Table \ref{tab:comp_to_state_of_art_hinter_f1_scores} for the dataset of Hinterstoisser \etal ~\cite{hinterstoisser2013model}, our approach with the iterative process improves the F1-Score by 2.8\% on average and up to 6.4\% on some objects. The biggest gains are seen in objects that have large amounts of indistinct regions (image patches) for which background clutter can easily be confused. For example, the biggest improvements are seen in the Glue, Holepuncher and Phone objects that contain large planar regions. Furthermore, in Table \ref{tab:comp_to_state_of_art_hinter_f1_scores} we compare the results of using holistic LINEMOD templates, to scale-invariant (SI) LINEMOD templates. As the scale-invariant version is trained using only one scale, the performance is increased 6-fold (623 templates opposed to 3738). Furthermore, the performance is also increased by 6.9\% on average, this is due to the fact that templates are able to matched at scales not seen in the template learning stage of the original LINEMOD. The impact of our iterative method is more visible in testing scenes that contain small objects with significant amount of foreground occlusions, e.g. Camera object in \textit{Domestic Environment Dataset} or Coffee Cup object in \textit{Bin-Picking Dataset.}   

Additionally, we investigate the impact of different modalities on the performance of our system. We used a much smaller validation subset  comprised of images of two of the objects contained in both our \textit{Domestic Environment Dataset} and \textit{Bin-Picking Dataset}. Table \ref{tab:comp_modal} summarizes the calculated F1-Scores, while one interesting finding is that geometrical cues seem to be more important from appearance ones when trying to detect the object Coffee Cup. In turn, although the detection levels are relatively low, the Camera object favors the usage of both RGB and D modalities.  

\subsubsection{$1^{st}$ Dataset: LINEMOD}\label{sec:comp_LINEMOD}
The performance of our method was evaluated also on the famous dataset of of Hinterstoisser \etal ~\cite{hinterstoisser2013model} against the state of the art works of LINEMOD \cite{hinterstoisser2011multimodal}, Drost \etal ~\cite{drost2010model}, Rios Cabrera \etal ~\cite{Rios_Cabrera_2013_ICCV} and Brachmann \etal ~\cite{brachmann2014learning}. This dataset contains only one object instance per testing image in a cluttered background, however, without occlusions. Table \ref{tab:comp_to_state_of_art_hinter_percentages} summarizes the results of the conducted comparison with respect to the performance levels and the execution times. In the second column of Table \ref{tab:comp_to_state_of_art_hinter_percentages} we report which matching score (Eq. \ref{non_sym_eq} or Eq. \ref{sym_eq}) was used per object, while for our method we used the same $k_m$ threshold values with the ones presented in ~\cite{hinterstoisser2013model}. The performance of our method is almost identical to the one of Brachmann \etal ~\cite{brachmann2014learning} on a dataset that could be apprehended as obsolete, since the challenges it offers, i.e. background clutter with no foreground occlusions, are experimentally proven to be easily addressed. 
As far as execution times are concerned, the scalable method of Rios Cabrera \etal ~\cite{Rios_Cabrera_2013_ICCV} provided the lowest runtimes followed by LINEMOD \cite{hinterstoisser2011multimodal} and our method, respectively. Compared to the aforementioned works, our method provides higher performance levels with the additional computational cost. Unfortunately, we could not identify reported runtimes for the method of Brachmann \etal ~\cite{brachmann2014learning}. We would like to note that since the evaluation metric for symmetric objects is not very representative, one could transform every hypothesis using the rotation matrices $R_x$ that convert the object to its symmetric shapes and evaluate the detection performance using the non-symmetric measure of Eq. \ref{non_sym_eq}.    

\emph{Implementation details}: We have implemented our own version of LINEMOD that produced the same exactly results presented in \cite{hinterstoisser2013model}. For the method of Drost \etal ~\cite{drost2010model}, we use a binary version kindly provided by the author and set the parameters as described in \cite{drost2010model}. Source code of the method of Brachmann \etal ~\cite{brachmann2014learning} is publicly available. As far as the method of Rios Cabrera \etal ~\cite{Rios_Cabrera_2013_ICCV} is concerned, we show in Table \ref{tab:comp_to_state_of_art_hinter_percentages} the respective performance levels as they were presented in the original paper. Regarding runtimes, reported numbers are taken from \cite{Rios_Cabrera_2013_ICCV}.

\begin{table}[h]
\small
\centering
\caption{Comparison of two versions of our \textit{Latent-Class Hough Forests} and the method of Brachmann \etal~\cite{brachmann2014learning}.}
\label{tab:our_dome}
\begin{tabular}{lcccc}
\multicolumn{4}{c}{\textbf{Domestic Environment Dataset}} \\
& & \multicolumn{2}{c}{\textit{Latent-Class Hough Forests}} \\ \hline
\multicolumn{1}{l|}{\textbf{Object}}	&\multicolumn{1}{l|}{\textbf{~\cite{brachmann2014learning}}}	&\multicolumn{1}{l|}{\textbf{5 trees 1/2 patch}}	&\multicolumn{1}{l}{\textbf{10 trees 2/3 patch}} \\ \hline
\multicolumn{1}{l|}{Coffee Cup}		&91.2\%											&92.4\%								&\textbf{94.6\%}		\\ 
\multicolumn{1}{l|}{Shampoo}		&82.4\%											&86.9\%								&\textbf{88.2\%}		\\
\multicolumn{1}{l|}{Joystick}		&\textbf{75.9\%}									&71.3\%								&74.1\%				\\
\multicolumn{1}{l|}{Camera}		&69.1\%											&73.8\%								&\textbf{78.4\%}		\\
\multicolumn{1}{l|}{Juice Carton}	&89.7\%											&91.3\%								&\textbf{93.5\%}		\\
\multicolumn{1}{l|}{Milk}			&47.6\% 											&50.1\% 								&\textbf{51.6\%}		\\ \hline
\multicolumn{1}{l|}{Average}		&75.9\% 											&77.6\%								&\textbf{80.0\%}		\\               
\end{tabular}
\end{table}

\subsubsection{$2^{nd}$ Dataset: Domestic Environments}\label{sec:comp_Domestic}

\begin{figure*}[!]
	\begin{center}
			{\includegraphics[width=0.24\textwidth]{images/color394.jpg}}
			{\includegraphics[width=0.24\textwidth]{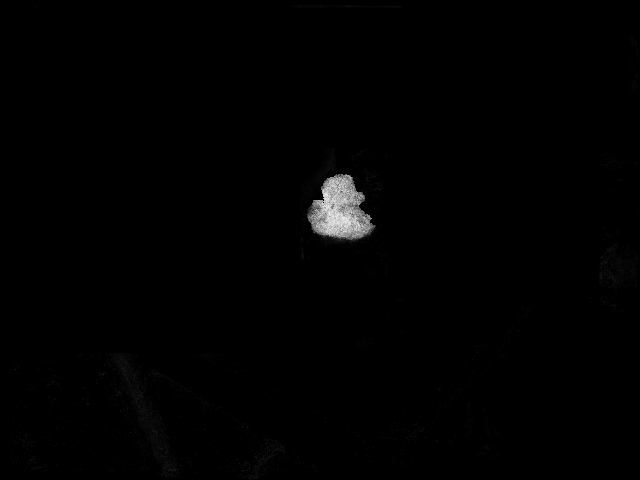}}
			{\includegraphics[width=0.24\textwidth]{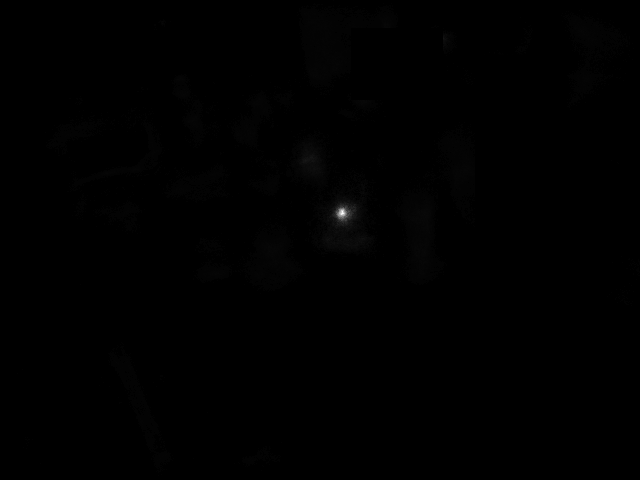}}
			{\includegraphics[width=0.24\textwidth]{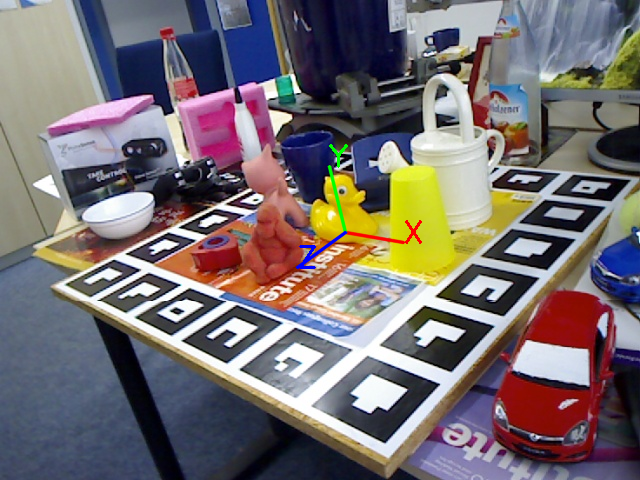}}
			{\includegraphics[width=0.24\textwidth]{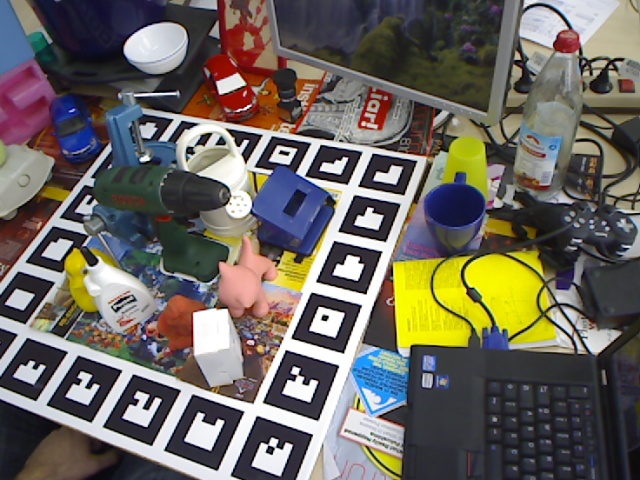}}
			{\includegraphics[width=0.24\textwidth]{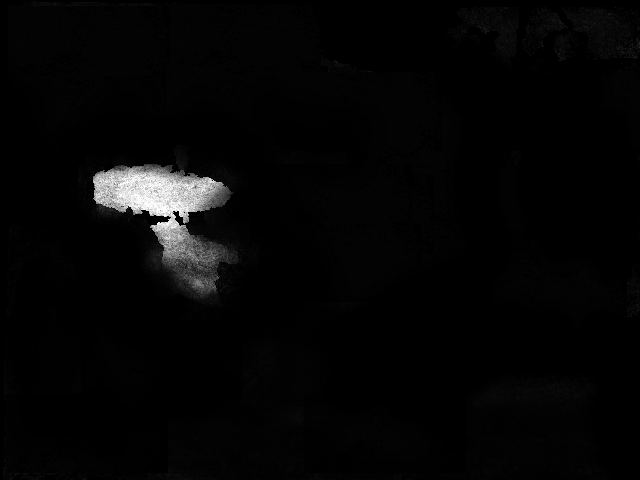}}
			{\includegraphics[width=0.24\textwidth]{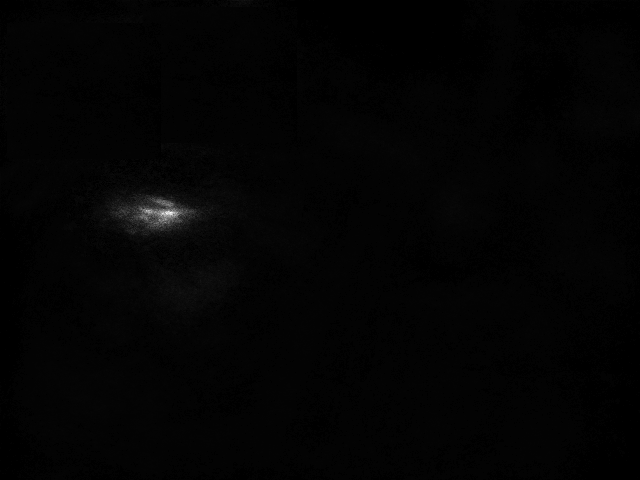}}
			{\includegraphics[width=0.24\textwidth]{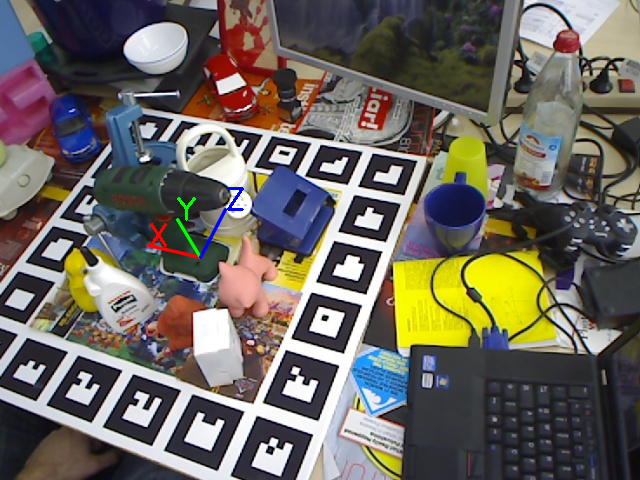}}
			{\includegraphics[width=0.24\textwidth]{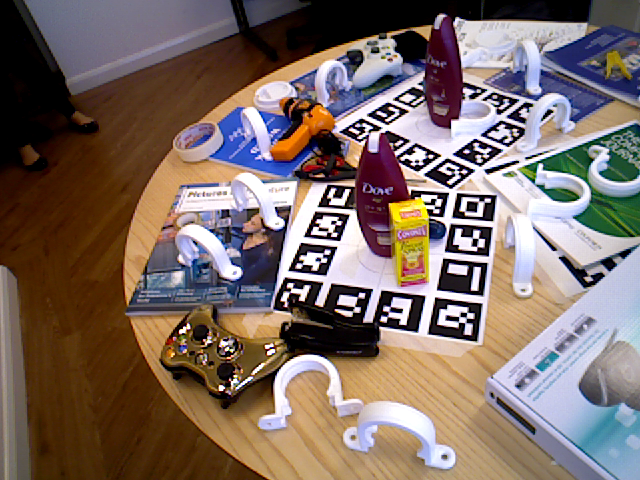}}
			{\includegraphics[width=0.24\textwidth]{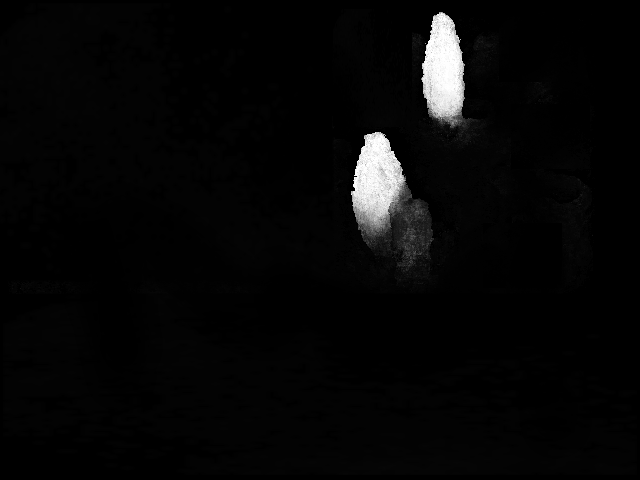}}
			{\includegraphics[width=0.24\textwidth]{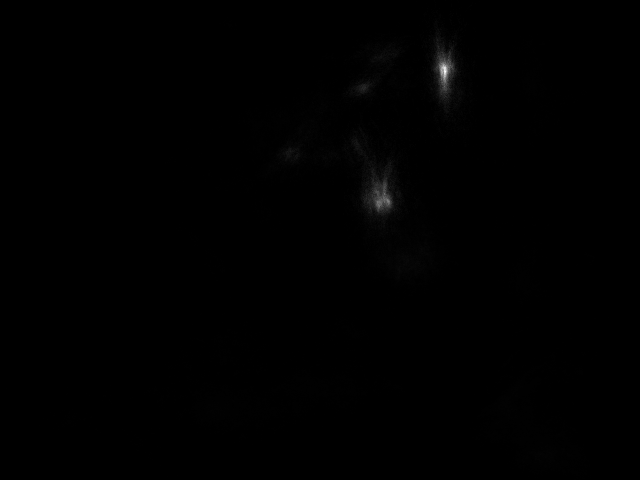}}
			{\includegraphics[width=0.24\textwidth]{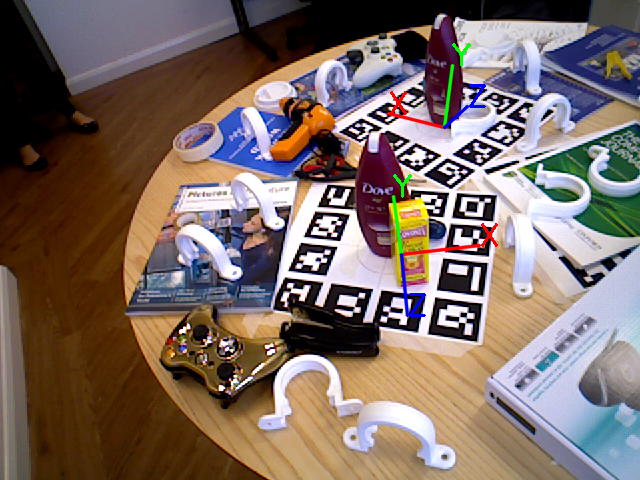}}
			{\includegraphics[width=0.24\textwidth]{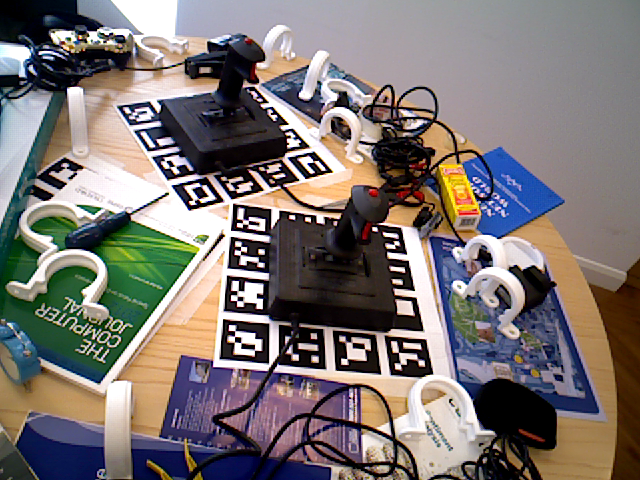}}
			{\includegraphics[width=0.24\textwidth]{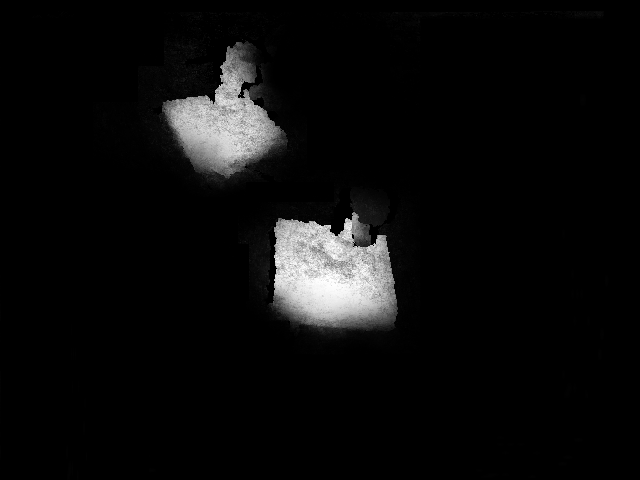}}
			{\includegraphics[width=0.24\textwidth]{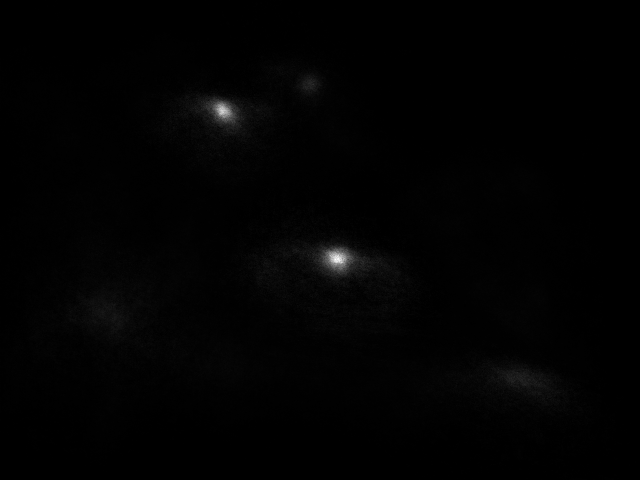}}
			{\includegraphics[width=0.24\textwidth]{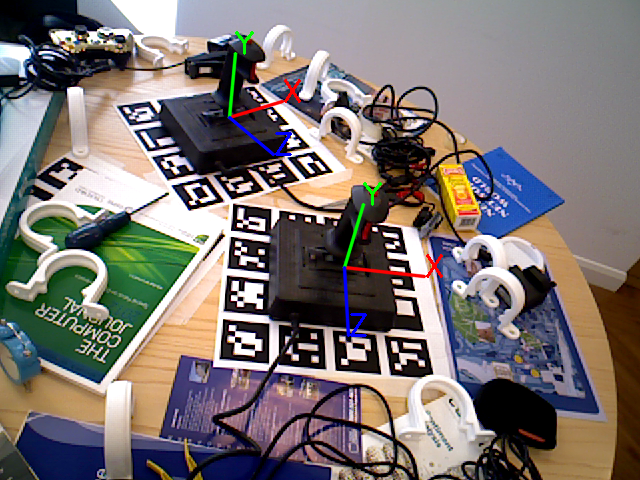}}
			{\includegraphics[width=0.24\textwidth]{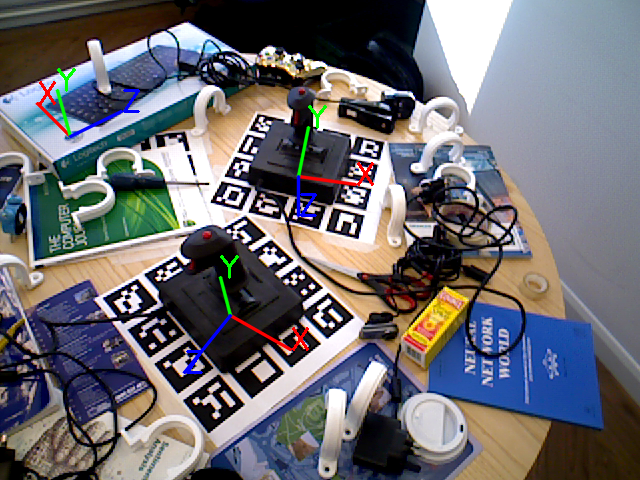}}
			{\includegraphics[width=0.24\textwidth]{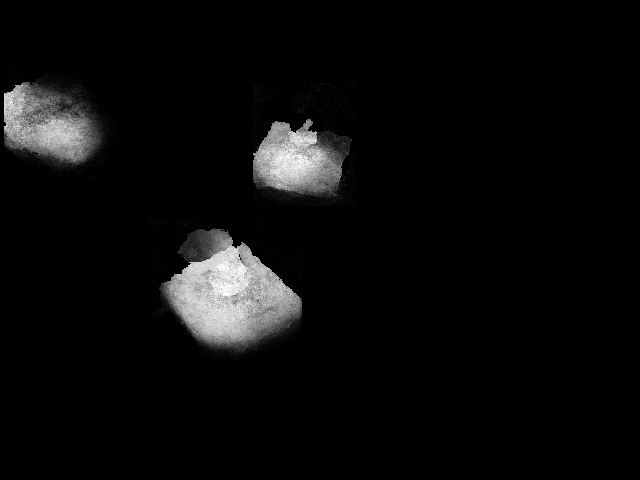}}
			{\includegraphics[width=0.24\textwidth]{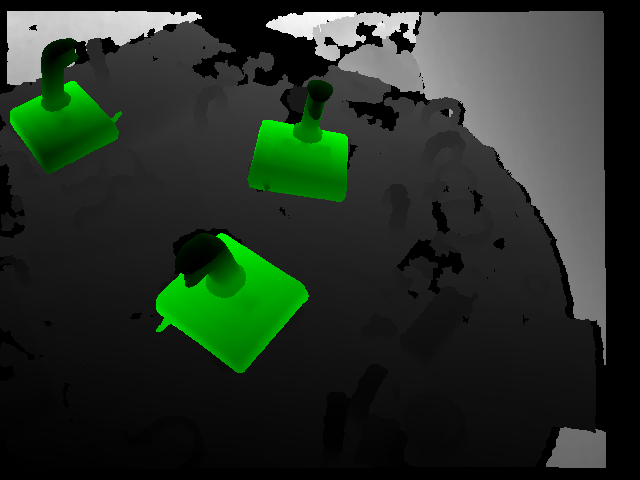}}
			{\includegraphics[width=0.24\textwidth]{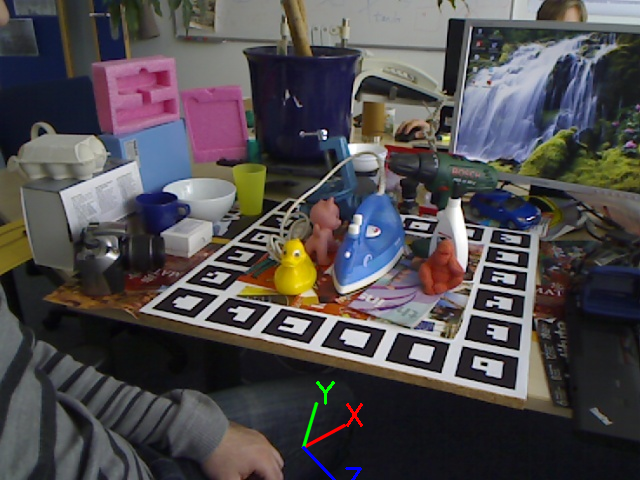}}
\end{center}
\caption{Rows 1-4 show, from left to right, the original RGB image, the final segmentation mask, the final Hough vote map and the augmented 3D axis of the estimated result. The final row shows some incorrect results, from left to right: one false positive leading to a false segmentation mask and wrong 3D rendering and finally a false negative.}\label{fig:qual_res}
\end{figure*}

Our \textit{Domestic Environment Dataset} consists of 6 objects placed on top of a table simulating, thus, a kitchen table in a house. We provide a dense 3D reconstruction of each object obtained via a commercially available 3D scanning tool~\cite{Doe:2009:Misc}. Additionally, we provide for each object, similarly to \cite{hinterstoisser2013model}, an individual testing sequence containing over 700 images annotated with ground truth position and 3D pose. Testing sequences were obtained by a freely moving handheld RGB-D camera and ground truth was calculated using marker boards and verified manually. The testing images were sampled to produce sequences that are uniformly distributed in the pose space by $\left[0^{\circ}-360^{\circ}\right]$, $\left[-80^{\circ}-80^{\circ}\right]$ and $\left[-70^{\circ}-70^{\circ}\right]$ in the yaw, roll and pitch angles, respectively. Unlike the dataset of \cite{hinterstoisser2013model}, our testing sequences contain \emph{multiple object instances} and \emph{foreground occlusions} along with near and far range 2D and 3D clutter, making it more challenging for the task of 3D object detection and pose estimation. Some example frames from this dataset can be seen in Fig \ref{fig:qual_res}. 

In Fig. \ref{fig:comp_to_state_of_arts} we show the average precision-recall curves across all objects and in Table \ref{tab:comp_to_state_of_art_hinter_f1_scores} we show the F1-Score per object for each dataset. All methods provided evidence of worse performance levels on the new dataset, which is to be suspected due to the introduction of occlusions as well as multiple object instances. As can be seen we outperform both state of the arts in both datasets. The method of Drost \etal ~\cite{drost2010model} has considerably lower precision values due to the fact that it does not take object boundaries into consideration, thus large planar regions of the target object can have a large surface overlap in the background clutter causing many false positives in addition to the true positives. Conversely, our method maintains high levels of precision at high recall which is due to the inference process simplifying the Hough space. 


We have also compared our method with the one of Brachmann \etal ~\cite{brachmann2014learning} that is designed to estimate only one object per image. For a fair comparison, we compare the scores of the top hypothesis produced by the respective methods per image. Table \ref{tab:our_dome} shows the detection results for the method of Brachmann \etal ~\cite{brachmann2014learning} and two different versions of our \textit{Latent-Class Hough Forests}. In Fig. \ref{fig:qual_res} we present some qualitative results on both our \textit{Domestic Environment Dataset} and the one of LINEMOD \cite{hinterstoisser2013model}. A video demonstrating the efficiency of our method in this dataset is also available\footnote{\url{https://www.youtube.com/watch?v=idY3Q7wg5rk}}.

\subsubsection{$3^{rd}$ Dataset: Bin-picking}\label{sec:comp_Bin_Picking}
One of the most widely encountered application in industry is the one of robotic manipulation and, specifically, the one of manipulating similar objects placed in bins (e.g. grasping bolts and screws from a box). It is apparent that, this particular working scenario is very challenging for any vision algorithm since several cascading issues arise (e.g. severe occlusions, foreground clutter etc.). Annotating objects stacked in bins can not be done via the methodology we used for our \textit{Domestic Environments Dataset}, where we placed markers beneath each testing object in order to acquire the necessary ground truth measurements. Since there is no simple way of obtaining the latter, Liu \etal ~\cite{liu2012} devised a statistical evaluation of the pose estimation consistency across multiple viewpoints of the camera. They utilized a robot arm to register the different viewpoints, while the final metric is a histogram of the deviations from the median pose estimation. 

Unilke \cite{liu2012}, we put manual labor to provide, to the best of our knowledge, the first fully annotated \textit{Bin-picking Dataset}. The building of the dataset was divided into two phases, firstly, registering the viewpoints and secondly, annotating the objets. A freely moving handheld RGB-D camera was used to capture the testing frames, which, in turn, are registered by manually selecting key-point correspondences across all sequences. Afterwards, we annotate each object in the testing scenes by manually projecting the 3D mesh of the targets onto the respective point clouds of the testing sequences. The dataset comprises of multiple instances of two objects (Juice Carton and Coffee Cup) in more than 150 testing images. 

\begin{figure*}[!]
	\begin{center}
			{\includegraphics[width=0.24\textwidth]{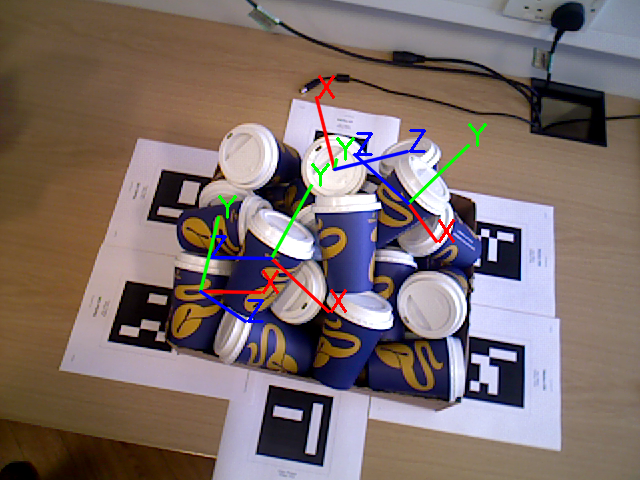}}
			{\includegraphics[width=0.24\textwidth]{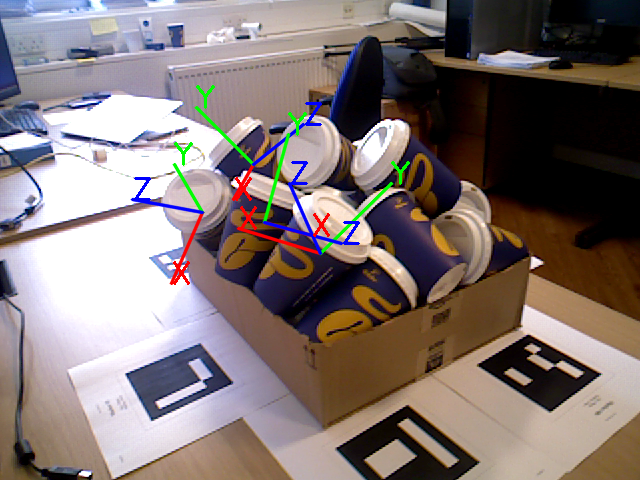}}						
			{\includegraphics[width=0.24\textwidth]{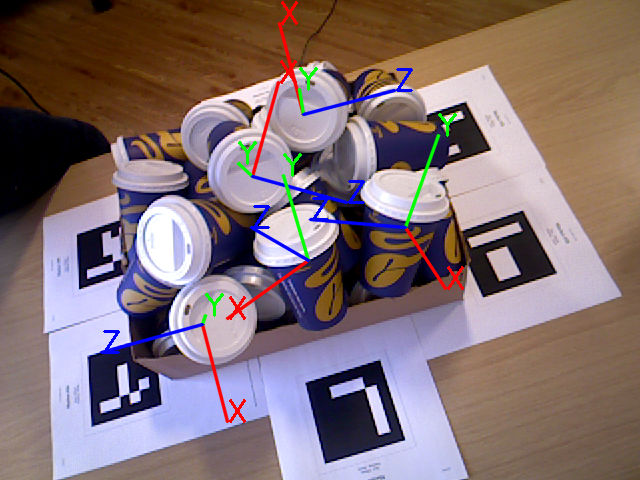}}
			{\includegraphics[width=0.24\textwidth]{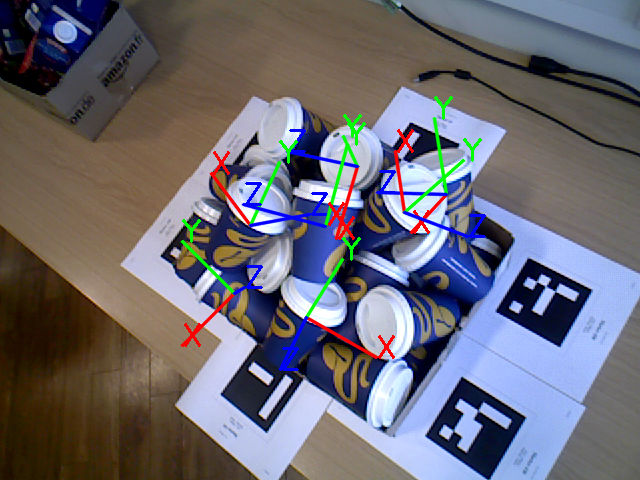}}		
			{\includegraphics[width=0.24\textwidth]{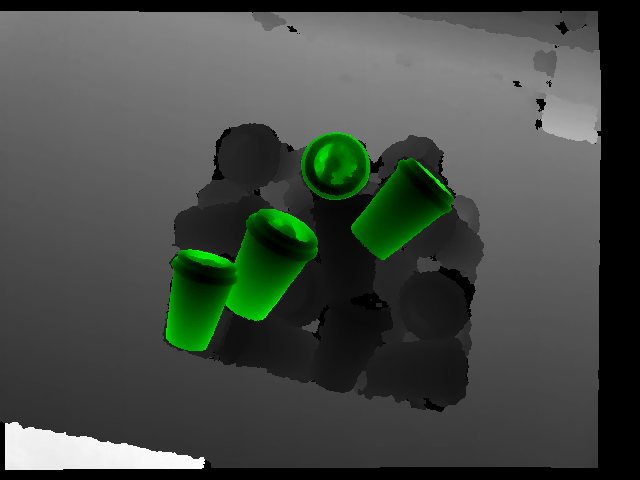}}
			{\includegraphics[width=0.24\textwidth]{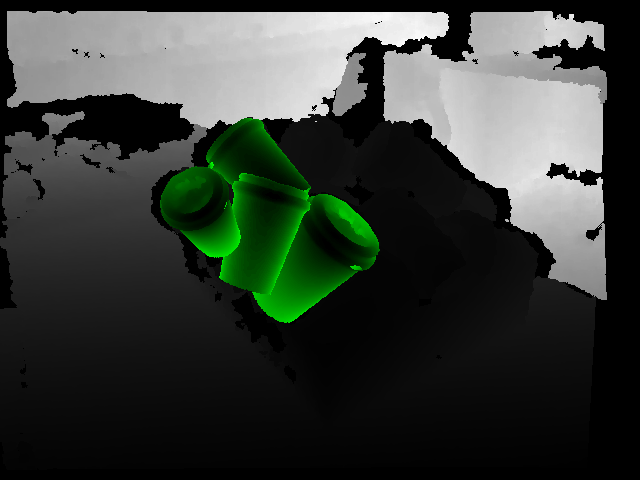}}						
			{\includegraphics[width=0.24\textwidth]{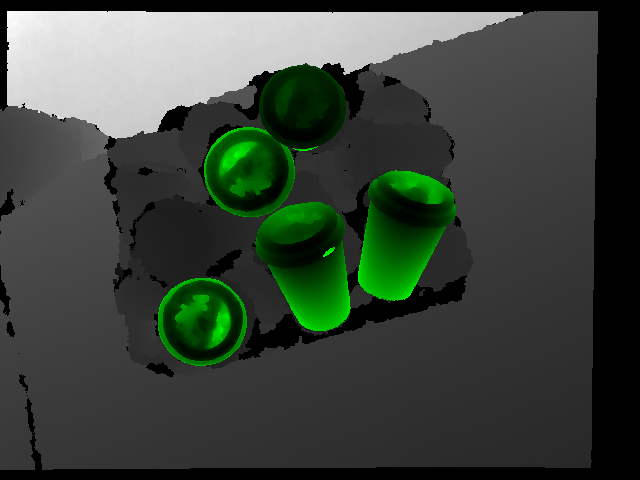}}
			{\includegraphics[width=0.24\textwidth]{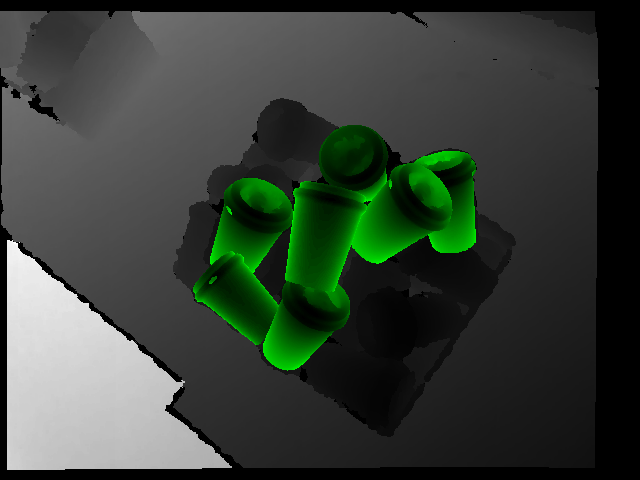}}	
			{\includegraphics[width=0.24\textwidth]{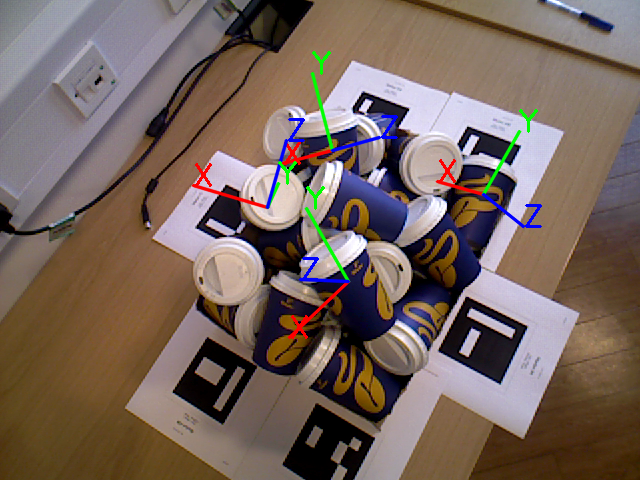}}
			{\includegraphics[width=0.24\textwidth]{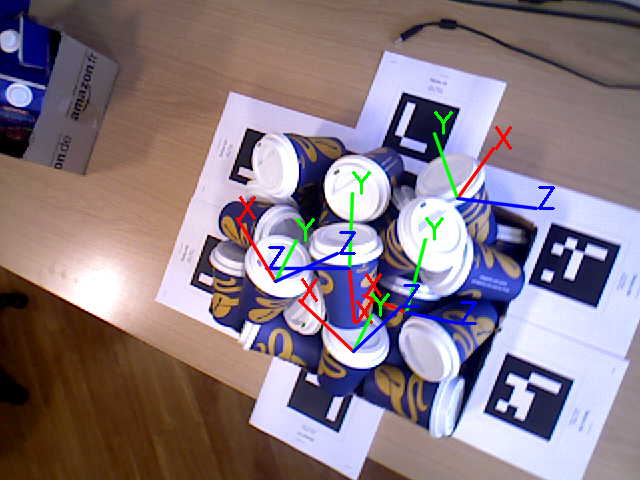}}
			{\includegraphics[width=0.24\textwidth]{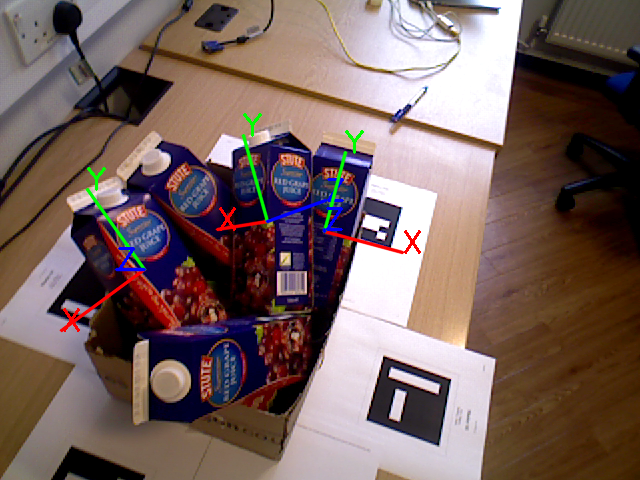}}
			{\includegraphics[width=0.24\textwidth]{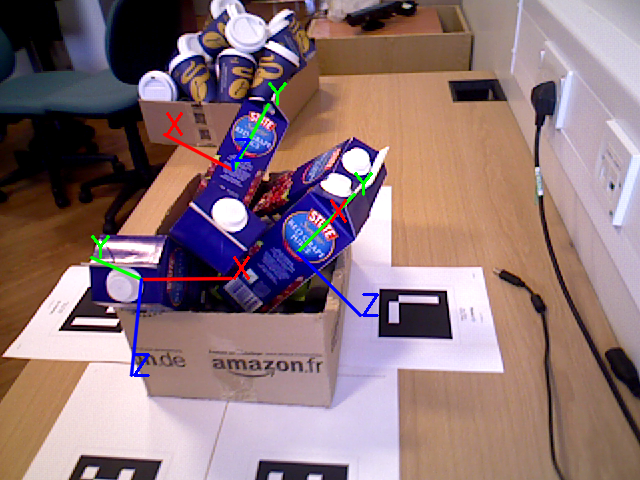}}
			{\includegraphics[width=0.24\textwidth]{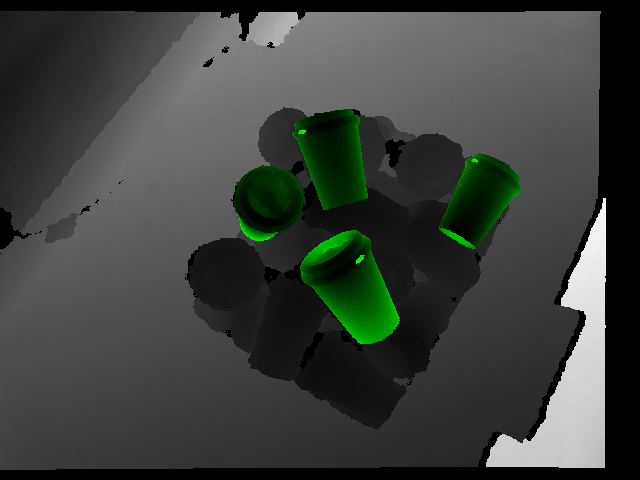}}	
			{\includegraphics[width=0.24\textwidth]{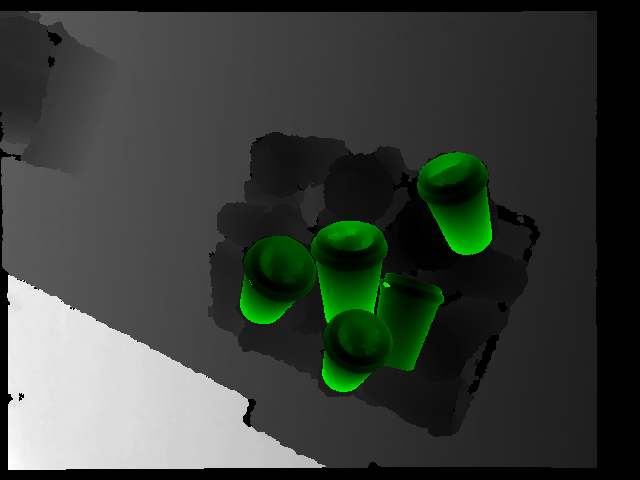}}
			{\includegraphics[width=0.24\textwidth]{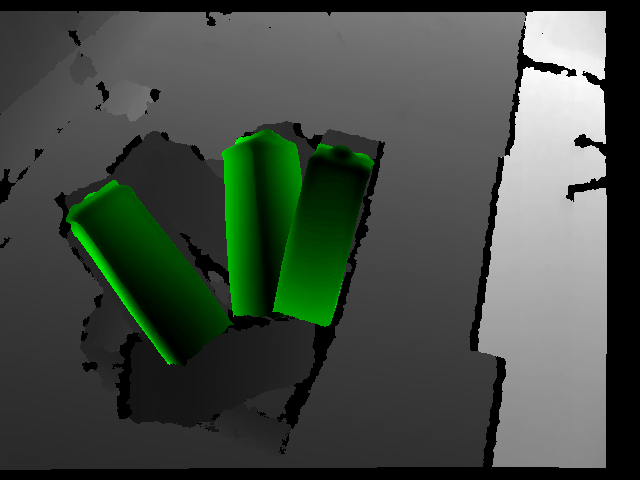}}
			{\includegraphics[width=0.24\textwidth]{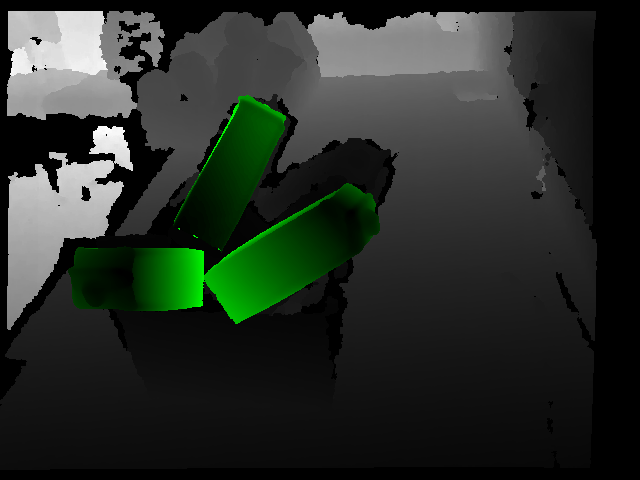}}
			{\includegraphics[width=0.24\textwidth]{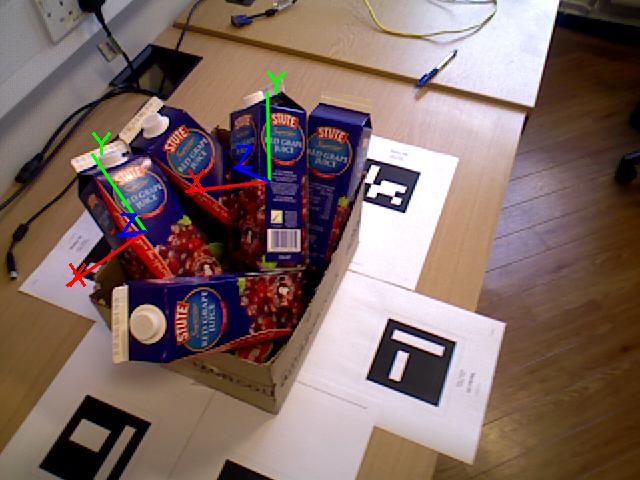}}
			{\includegraphics[width=0.24\textwidth]{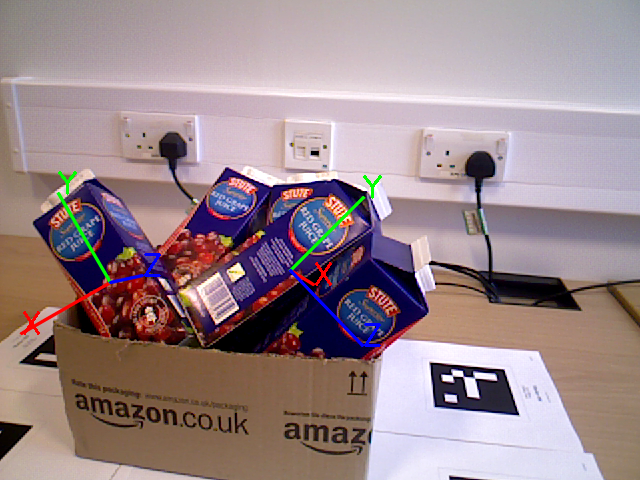}}
			{\includegraphics[width=0.24\textwidth]{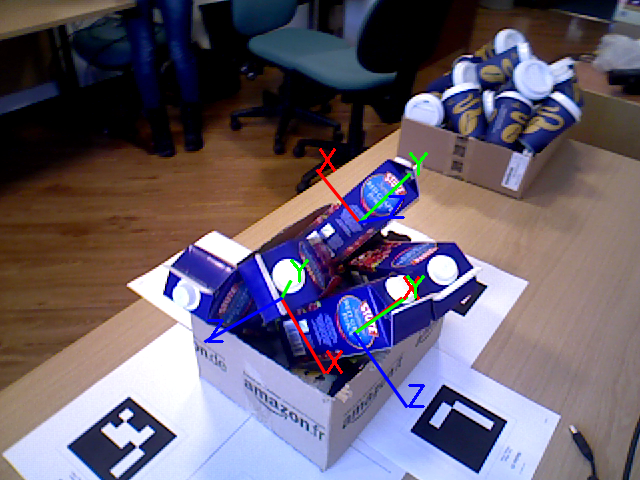}}
			{\includegraphics[width=0.24\textwidth]{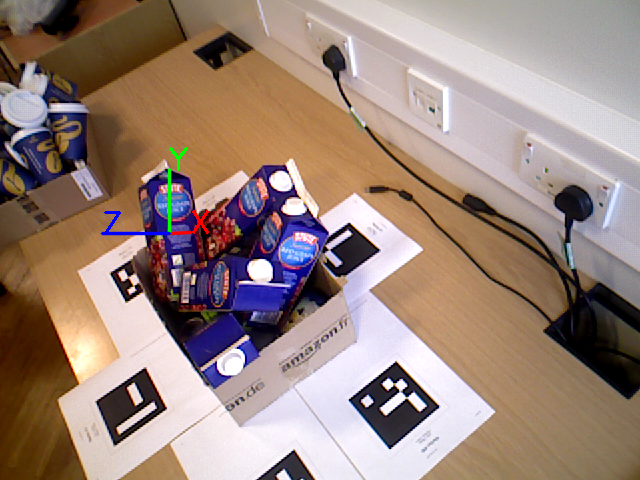}}
			{\includegraphics[width=0.24\textwidth]{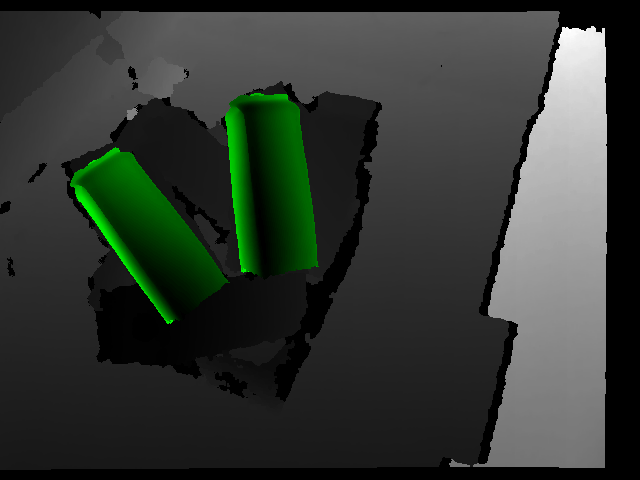}}
			{\includegraphics[width=0.24\textwidth]{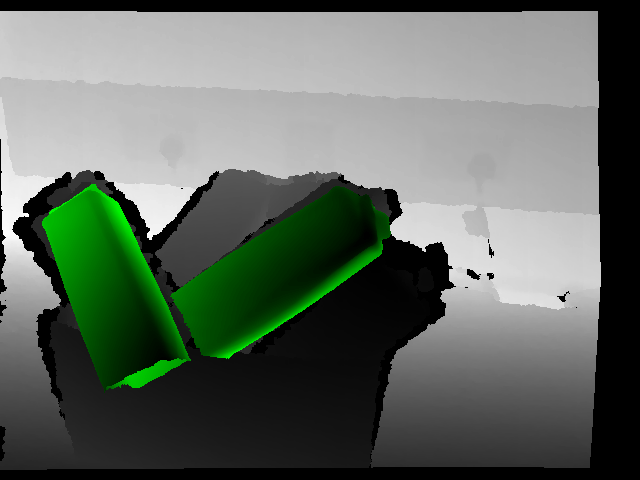}}
			{\includegraphics[width=0.24\textwidth]{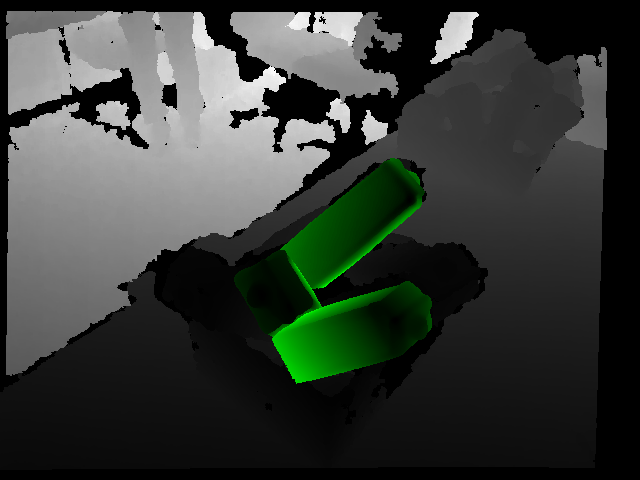}}
			{\includegraphics[width=0.24\textwidth]{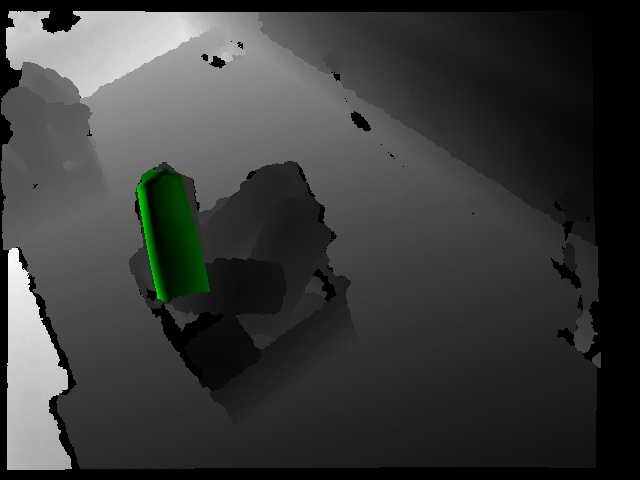}}
			{\includegraphics[width=0.24\textwidth]{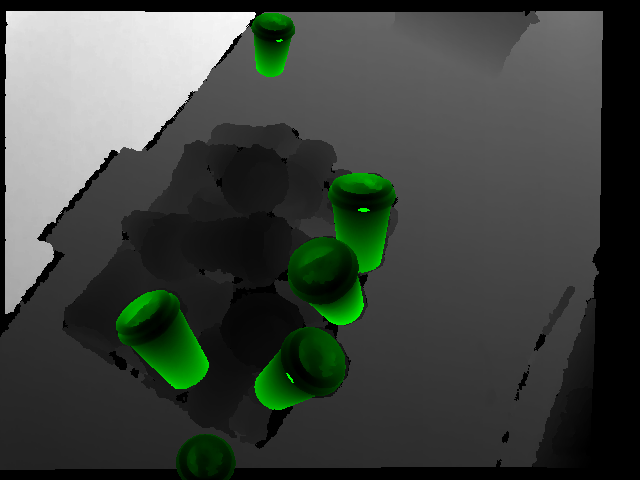}}	
			{\includegraphics[width=0.24\textwidth]{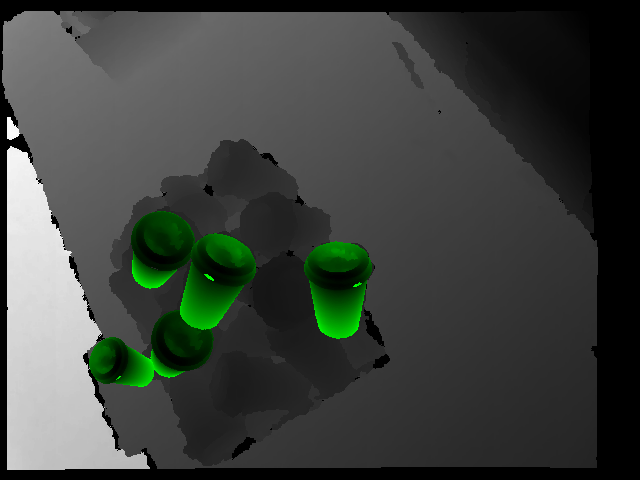}}	
			{\includegraphics[width=0.24\textwidth]{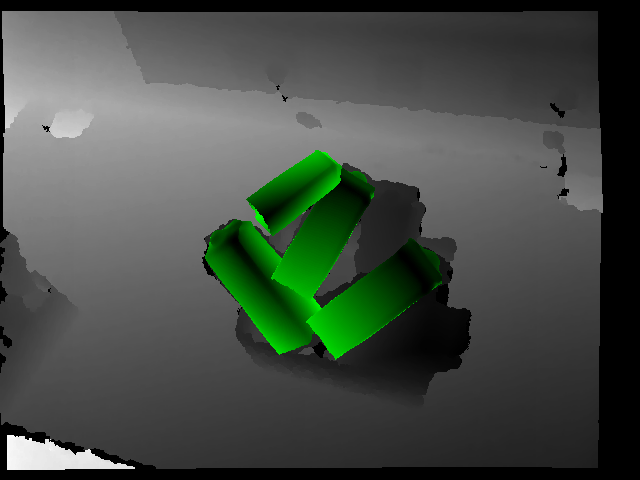}}	
			{\includegraphics[width=0.24\textwidth]{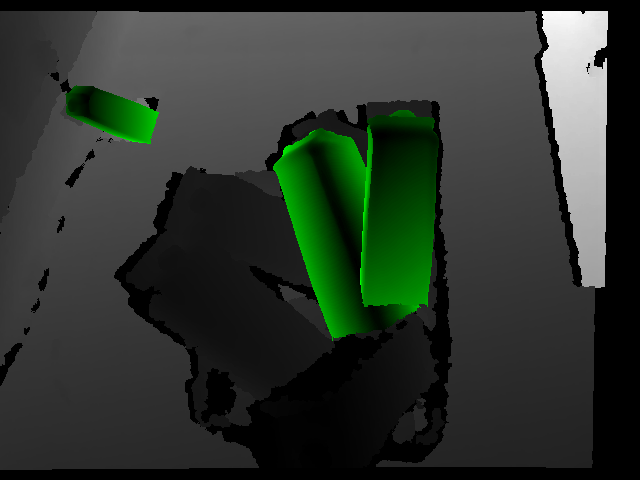}}
\end{center}
\caption{Rows 1, 3 and 5 show the augmented 3D axis of the estimated result for our \textit{Bin-picking Dataset}. Rows 2, 4 and 6 illustrate the overlaid mesh of the registered objects for the scene right above. The final row shows cases of false positives.}\label{fig:qual_res_bin}
\end{figure*}

In Table \ref{tab:our_binpick} we show the efficiency of our method in this challenging dataset in the form of accurate detection and F1-Scores. Similar to our previous comparison with the method of Brachmann \etal ~\cite{brachmann2014learning}, we evaluate the scores of the top hypothesis produced by the respective methods per image. F1-Scores are calculated only for our method which is designed to work with multi-instance objects. In Fig. \ref{fig:qual_res_bin} we present several qualitative results on the \textit{Bin-picking Dataset}. Our system is capable of achieving near real-time execution without any GPU processing. A video demonstrating the efficiency of our method in the \textit{Bin-picking Dataset} is also available\footnote{\url{https://www.youtube.com/watch?v=dh2VtnnsGuY}}.

\begin{table}[h]
\small
\centering
\caption{Percentages of accurate detections and F1-Scores for two versions of our \textit{Latent-Class Hough Forests} and the method of Brachmann \etal~\cite{brachmann2014learning}.}
\label{tab:our_binpick}
\begin{tabular}{lcccc}
\multicolumn{4}{c}{\textbf{Bin-Picking Dataset}} \\
& & \multicolumn{2}{c}{\textit{Latent-Class Hough Forests}} \\ \hline
\multicolumn{1}{l|}{\textbf{Object}}	&\multicolumn{1}{l|}{\textbf{~\cite{brachmann2014learning}}}	&\multicolumn{1}{l|}{\textbf{5 trees 1/2 patch}}	&\multicolumn{1}{l}{\textbf{10 trees 2/3 patch}} \\ \hline
\multicolumn{1}{l|}{Coffee Cup}		&89.4\%											&90.1\% (0.521)						&\textbf{91.2\%} (0.542)	\\ 
\multicolumn{1}{l|}{Juice Carton}	&87.6\%											&89.6\% (0.484)						&\textbf{90.4\%} (0.492)	\\ \hline
\multicolumn{1}{l|}{Average}		&88.5\% 											&89.8\% (0.502)						&\textbf{90.8\%} (0.517)	\\       
\end{tabular}
\end{table}

\section{Conclusion}
\label{sec:conclusion}
In this paper we have introduced a novel framework for accurate 3D detection and pose estimation of multiple object instances in cluttered and occluded scenes. We have demonstrated that these challenges can be efficiently met via the adoption of a state of the art template matching feature into a patch-based regression forest. During training we employ a one-class learning scheme, i.e. training with positive samples only rather than involving negative examples. In turn, during inference, we engage the proposed \textit{Latent-Class Hough Forests} that iteratively produce a more accurate estimation of the clutter / occluder distribution by considering class distribution as latent variables. As a result, apart from accurate detection results we can, further, obtain an highly representative occlusion-aware masks facilitating further tasks such as scene layout understanding, occlusion aware ICP or online domain adaption to name a few. Our method is evaluated using both the public dataset of Hinterstoisser \etal~\cite{hinterstoisser2013model} and our new challenging ones containing foreground occlusion (severe in cases of the \textit{Bin-Picking Dataset}) and multiple object instances. Experimental evaluation provides evidence of our novel \textit{Latent-Class Hough Forest} outperforming all baselines highlighting the potential benefits of part-based strategies to address the issues of such a challenging problem. 


\bibliographystyle{IEEEtranS.bst}
\bibliography{main.bib}
\vspace{-1cm}

\begin{IEEEbiography}[{\includegraphics[width=1in,height=1.25in,clip,keepaspectratio]{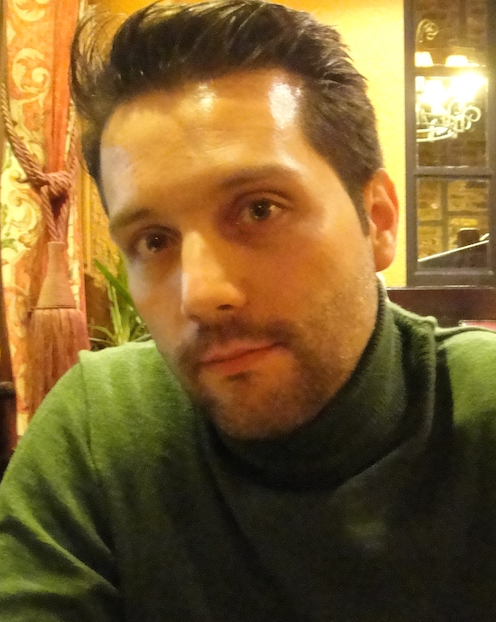}}]{Rigas Kouskouridas}received his Ph.D. degree from the Department of Production and Management Engineering of the Democritus University of Thrace in 2013. He is currently a Postdoctoral research associate in the Imperial Computer Vision and Learning Lab of the Electrical and Electronic Engineering Department of Imperial College London. His areas of interest include computer vision, machine learning and robotics.
\end{IEEEbiography}
\vspace{-1.5cm}
\begin{IEEEbiography}[{\includegraphics[width=1in,height=1.25in,clip,keepaspectratio]{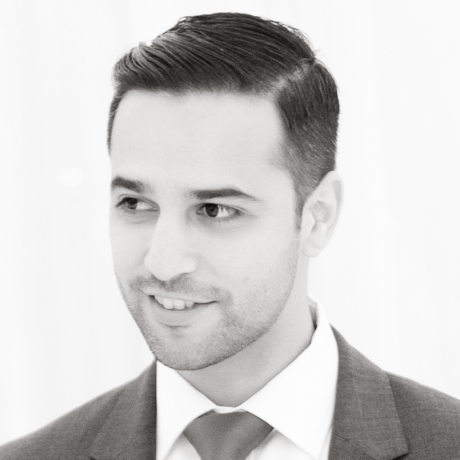}}]{Alykhan Tejani}received his MEng degree in Computing in 2010 and his MPhil whilst at the Computer Vision and Learning Lab in 2014, both at Imperial College London. He currently works at Blippar developing their visual search and augmented reality platforms. His current research interests include large-scale object recognition and 3D object pose estimation.
\end{IEEEbiography}
\vspace{-1.6cm}
\begin{IEEEbiography}[{\includegraphics[width=1in,height=1.25in,clip,keepaspectratio]{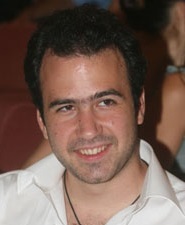}}]{Andreas Doumanoglou}received his MEng in Electrical and Electronic Engineering from the Aristotle University of Thessaloniki in 2009. Currently, he is Ph.D candidate in the Imperial Computer Vision and Learning Lab of the Electrical and Electronic Engineering Department of Imperial College London. He works in collaboration with CERTH-ITI and his areas of interest include robot vision and machine learning.
\end{IEEEbiography}
\vspace{-1.5cm}
\begin{IEEEbiography}[{\includegraphics[width=1in,height=1.25in,clip,keepaspectratio]{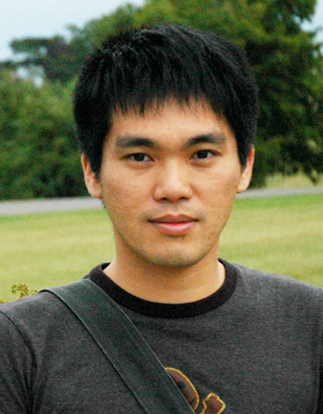}}]{Danhang Tang}is a PhD candidate in the Imperial Computer Vision and Learning Lab of the Electrical and Electronic Engineering Department of Imperial College London. Prior to his PhD, he received a 1st honor MSc degree from University College London and a BSc degree from Sun Yat-sen University. From 2007 to 2009, he worked as system architect for Evryx Technologies Ltd., in support for SnapNow, one of the first image recognition apps in the world. 
\end{IEEEbiography}
\vspace{-1.5cm}
\begin{IEEEbiography}[{\includegraphics[width=1in,height=1.25in,clip,keepaspectratio]{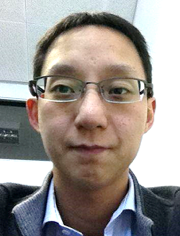}}]{Tae-Kyun (T-K) Kim}is an Assistant Professor and leader of Computer Vision and Learning Lab at Imperial College London, UK, since Nov 2010. He received the B.Sc. and M.Sc. degrees from Korea Advanced Institute of Science and Technology in 1998 and 2000, respectively, and worked at Samsung Advanced Institute of Technology in 2000-2004. He obtained his PhD from Univ. of Cambridge in 2008 and Junior Research Fellowship (governing body) of Sidney Sussex College, Univ. of Cambridge for 2007-2010. His research interests primarily lie in decision forests (tree-structure classifiers) and linear methods for: articulated hand pose estimation, face analysis and recognition by image sets and videos, 6D object pose estimation, active robot vision, activity recognition and object detection/tracking. He has co-authored over 40 academic papers in top-tier conferences and journals in the field, his co-authored algorithm for face image retrieval is an international standard of MPEG-7 ISO/IEC. 
\end{IEEEbiography}
\end{document}